\documentclass[preprint]{elsarticle}

\usepackage[utf8]{inputenc}
\usepackage[frozencache, cachedir=.]{minted}
\usepackage{subcaption}
\usepackage{graphicx}
\usepackage{wrapfig}
\usepackage{tabularx}
\usepackage{multirow}
\usepackage{adjustbox}
\usepackage{arydshln}

\usepackage{color}
\usepackage{float}
\usepackage{booktabs}
\usepackage{mdframed}
\usepackage[thmmarks]{ntheorem}
\usepackage[section]{placeins}
\usepackage{problogconfig}
\usepackage{textcomp}
\usepackage[inline]{enumitem}

\usepackage{amsmath}
\usepackage{amssymb}

\newcommand{\h}{\boldsymbol{h}}
\newcommand{\x}{\boldsymbol{x}}
\newcommand{\y}{\boldsymbol{y}}

\theoremstyle{break}
\theorembodyfont{\normalfont}
\newmdtheoremenv[topline=false,rightline=false,bottomline=false]{example}{Example}
 
\theoremstyle{break}
\theorembodyfont{\normalfont}
\theoremsymbol{\ensuremath{\triangleleft}}
\newtheorem{definition}{Definition}

\newfloat{lstfloat}{htbp}{ext}
\floatname{lstfloat}{Listing}

\newcommand{\digit}[1]{\vcenter{\hbox{\includegraphics[height=10pt]{#1}}}}
\DeclareMathOperator*{\argmin}{arg\,min}

\DeclareMathOperator*{\lpif}{\mathtt{{:}{\---}}}
\DeclareMathOperator*{\prob}{\mathtt{{:}{:}}}
\date{}

\begin{document}

\begin{frontmatter}
\title{Neural Probabilistic Logic Programming in DeepProbLog\tnoteref{t1}}
\tnotetext[t1]{This is an extended and revised version of work previously published at NeurIPS 2018 \citep{deepproblog}.}

\author[kuleuven]{Robin Manhaeve\corref{mycorrespondingauthor}}
\cortext[mycorrespondingauthor]{Corresponding author}
\ead{robin.manhaeve@cs.kuleuven.be}
\author[kuleuven]{Sebastijan Duman\v{c}i\'c}
\author[cardiff]{Angelika Kimmig}
\author[ghentimec]{Thomas Demeester\fnref{shared}}
\author[kuleuven]{Luc De Raedt\fnref{shared}}

\address[kuleuven]{KU Leuven}
\address[cardiff]{Cardiff University}
\address[ghentimec]{Ghent University - imec}
\fntext[shared]{Joint last authors.}

\begin{abstract}
We introduce DeepProbLog, a neural probabilistic logic programming language that incorporates deep learning by means of neural predicates. We show how existing inference and learning techniques of the underlying probabilistic logic programming language ProbLog  can be adapted for the new language. We theoretically and experimentally demonstrate that DeepProbLog supports 
\begin{enumerate*}[label=(\roman*)]
\item both symbolic and subsymbolic representations and inference,
\item program induction,  
\item probabilistic (logic) programming, and
\item (deep) learning from examples. 
\end{enumerate*}
To the best of our knowledge, this work is the first to propose a framework where general-purpose neural networks and expressive probabilistic-logical modeling and reasoning are integrated in a way that exploits the full expressiveness and strengths of both worlds  and can be trained end-to-end based on examples.
\end{abstract}


\end{frontmatter}

\section{Introduction}
Many tasks in AI can be divided into roughly two categories: those that require low-level perception, and those that require high-level reasoning. At the same time, there is a growing consensus that being capable of tackling both types of tasks is essential to achieve true (artificial) intelligence \citep{kahneman}. 
Deep learning is empowering a new generation of intelligent systems that excel at low-level perception, where it is used  to interpret images, text and speech with unprecedented accuracy. The success of deep learning has caused a lot of excitement and has also created the impression that deep learning can solve any problem in artificial intelligence. 
However, there is a growing awareness of the limitations of deep learning: deep learning requires large amounts of (the right kind of) data to train the network, it provides neither justifications nor explanations, and the models are black-boxes that can neither be understood nor modified by domain experts. Although there have been attempts to demonstrate reasoning-like behaviour with deep learning \citep{santoro2017simple}, their current reasoning abilities are nowhere close to what is possible with typical high-level reasoning approaches.
The two most prominent frameworks for reasoning are logic and probability theory.  While in the past, these were studied by separate communities in artificial intelligence,  many researchers are working towards their integration, and  aim at combining probability with logic and 
statistical learning; cf. the areas of statistical relational artificial intelligence  \citep{DeRaedt16,Getoor07} and probabilistic logic programming \citep{deraedt15}.

The abilities of deep learning and statistical relational artificial intelligence approaches are complementary. While deep learning excels at low-level perception, probabilistic logics excel at  high-level reasoning. As such, an integration of the two would have very promising properties.
Recently, a number of researchers have revisited and modernized ideas originating from the field of neural-symbolic integration \citep{garcez2012neural}, searching for ways to combine the best of both worlds \citep{riedel2017programming,rocktaschel2017end,cohen2018tensorlog,santoro2017simple}, for example, by designing neural architectures representing differentiable counterparts of symbolic operations in classical reasoning tools. Yet, joining the full flexibility of high-level probabilistic and logical reasoning with the representational power of deep neural networks is still an open problem. 
Elsewhere \cite{nesy}, we have argued that neuro-symbolic integration should: 1) integrate neural networks with the two most prominent methods for reasoning, that is, logic and probability, and 2) that neuro-symbolic integrated methods should have the pure neural, logical and probabilistic methods as special cases.

With DeepProbLog, we tackle the neuro-symbolic challenge from this perspective. Furthermore, instead of integrating reasoning capabilities into a complex neural network architecture, we proceed the other way round. 
We start from an existing probabilistic logic programming language, ProbLog \citep{DeRaedt07}, and introduce the smallest extension that allows us to integrate neural networks: the neural predicate.
The idea is simple: in a probabilistic logic, atomic expressions of the form $q(t_1, ..., t_n)$ (aka tuples in a relational database) have a probability $p$. We extend this idea by allowing atomic expressions to be labeled with neural networks whose outputs can be considered probability distributions.
This simple idea is appealing as it allows us to retain all the essential components of the ProbLog language: the semantics, the inference mechanism, as well as the implementation.

Therefore,
{\bf \em one should not only integrate logic with neural networks in neuro-symbolic computation, but also probability}.

This effectively leads to an integration of probabilistic logics (hence statistical relational AI) with neural networks and opens up new abilities. Furthermore, 
although at first sight, this may appear as a complication, it  actually can greatly simplify the integration of neural networks with logic. 
The reason for this is that the probabilistic framework provides a clear optimisation criterion, namely the probability of the training examples. Real-valued probabilistic quantities are also well-suited for gradient-based training procedures, as opposed to discrete logic quantities.

\begin{example}
Before going into further detail, the following example illustrates the possibilities of this approach.
Consider the predicate $\mathtt{addition(X,Y,Z)}$, where $\mathtt{X}$ and $\mathtt{Y}$ are images of digits and $\mathtt{Z}$ is the natural number corresponding to the sum of these digits.
The goal is that after training, DeepProbLog allows us to make a probabilistic estimate on the validity of, for example, the example $\mathtt{addition(\digit{mnist_3},\digit{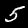},8)}$.
While such a predicate can be learned directly by a standard neural classifier, such an approach cannot incorporate background knowledge such as the definition of the addition of two {\em natural} numbers. 
In DeepProbLog such knowledge can easily be encoded in  rules such as \[\mathtt{addition(I_X,I_Y,N_Z)~{:}{-}~ digit(I_X,N_X), digit(I_Y,N_Y), N_Z~is~N_X + N_Y}\] with \texttt{is} the standard operator of logic programming to evaluate arithmetic expressions.
All that needs to be learned in this case is the neural predicate $digit$ which maps an image of a digit $I_D$ to the corresponding natural number $N_D$. The trained network can then be reused for arbitrary tasks involving digits.
Our experiments show that this leads not only to new capabilities but also to significant performance improvements.
An important advantage of this approach compared to standard image classification settings is that it can be  extended to multi-digit numbers without additional training.
We note that the single digit classifier (i.e., the neural predicate) is not explicitly trained by itself: its output can be considered a latent representation, as we only use training data with pairwise sums of digits.
\end{example}

To summarize, we introduce DeepProbLog which has a unique set of features: 
\begin{enumerate*}[label=(\roman*)]
\item
it is a programming language that supports neural networks and machine learning  and  has a well-defined semantics
\item it integrates logical reasoning with neural networks; so both symbolic and subsymbolic representations and inference;
\item it integrates probabilistic modeling, programming  and reasoning with neural networks (as DeepProbLog extends the probabilistic programming language ProbLog, which can be regarded as a very expressive directed graphical modeling language \citep{DeRaedt16});
\item it can be used to learn a wide range of probabilistic logical neural models from examples, including inductive programming. 
\end{enumerate*} 

This paper is a significantly extended and completed version of our previous work  \cite{deepproblog} (NeurIPS, spotlight presentation). This extended version now contains the necessary deep learning and  probabilistic logic programming background and a more in depth theoretical explanation. It also contains additional experiments (see Section~\ref{sec:experiments}): the MNIST addition experiments from the short version are completed with the new experiments \textbf{T3} and \textbf{T4}, 
and we designed new experiments (\textbf{T8} and \textbf{T9}) 
to further investigate the use of DeepProbLog on combined probabilistic learning and deep learning.
The code is available at \texttt{https://bitbucket.org/problog/deepproblog}.
\section{Background}
\subsection{Logic programming concepts}
In this section, we briefly summarize basic logic programming concepts; see e.g., \citet{lloyd:book89} for more details. 
Atoms are expressions of the form $q(t_1, ...,t_n)$ where $q$ is a predicate (of arity $n$, or $q/n$ in shorthand notation) and the $t_i$ are terms.
A literal is an atom or the negation $\neg q(t_1, ...,t_n)$ of an atom.
A term~$t$ is either a constant~$c$, a variable~$V$, or a structured term of the form $f(u_1, ...,u_k)$ where $f$ is a functor and the $u_i$ are terms. We  follow the Prolog convention and let constants, functors and predicates start  with a lower case character and variables with an upper case. 
A rule is an expression of the form $h \lpif ~b_1, ...,b_n$ where $h$ is an atom, the $b_i$ are literals, and all variables are universally quantified. Informally, the meaning of such a rule is that $h$ holds whenever the conjunction of the $b_i$ holds. 
Thus $\lpif$ represents logical implication ($\leftarrow$),  and the comma ($,$) represents conjunction ($\wedge$). 
Rules with an empty body $n=0$ are called facts. A logic program is a finite set of rules. 

A substitution $\theta = \{V_1 = t_1, ... , V_n = t_n\}$ is an assignment of terms~$t_i$ to variables~$V_i$. When applying a substitution $\theta$ to an expression $e$ we simultaneously replace all occurrences of $V_i$ by $t_i$ and denote the resulting expression as $e\theta$. Expressions that do not contain any variables are called ground.
The \emph{Herbrand base} of a logic program is the set of ground atoms that can be constructed using the predicates, functors and constants occurring in the program.\footnote{If the program does not contain constants, one arbitrary constant is added.} Subsets of the Herbrand base are called \emph{Herbrand interpretations}. A Herbrand interpretation is a \emph{model} of a clause $h \lpif b_1,\ldots ,b_n\ldotp$ if for every substitution $\theta$ such that 
the conjunction $(b_1,\ldots,b_n)\theta$ holds in the interpretation, $h\theta$ is in the interpretation. 

It is a model of a logic program if it is a model of all clauses in the program. 

For negation-free programs, the semantics is given by the minimal such model, known as the least Herbrand model, which is unique. General logic programs use the notion of negation as failure, that is, the negation of an atom is true exactly if the atom cannot be derived from the program. These programs are not guaranteed to have a unique minimal Herbrand model, and several ways to define a canonical model have been studied. We follow the well-founded semantics here \citep{VanGelder91}. 

The main inference task in logic programming is to determine whether a given atom $q$, also called \emph{query} (or \emph{goal}), is true in the canonical model of a logic program $P$, denoted by $P\models q$. If the answer is yes (or no), we also say that the query \emph{succeeds} (or \emph{fails}). If such a query is not ground, inference asks for the existence of an \emph{answer substitution}, that is, a substitution that grounds the query into an atom that is part of the canonical model. 

\subsection{Deep Learning}
 
The following paragraphs provide a very brief introduction to deep learning, focusing on concepts needed for understanding our work. Extensive further details can be found, e.g., in \cite{Goodfellow2016}. This section is meant to provide readers with little or no knowledge of deep learning, with a conceptual understanding of the main ideas.
In particular, we will focus on the setting of supervised learning, where the model learns to map an input item to a particular output, based on input-output examples.

An artificial neural network is a highly parameterized and therefore very flexible non-linear mathematical function that can be `trained' towards a particular desired behavior, by suitably adjusting its parameters. 

During training, the model learns to capture from the input data the most informative `features' for the task at hand. The need for `feature engineering' in classical (or rather, non-neural) machine learning methods has therefore been replaced by `architecture engineering', since a wide variety of neural network components are available to be composed into a suitable model.

Deep neural networks are often designed and trained in an `end-to-end' fashion, whereby only the raw input and the final target are known during training, and all components of the model are jointly trained.
For example, for the task of hand-written digit recognition, an input instance consists of a pixel image of a hand-written digit, whereas its target denotes the actual digit. 

Consider a supervised learning problem, with a training set $\{(\x_i, \y_i)\}_{i=1}^N$ containing $N$ i.i.d.~input instances $\x_i$ and corresponding outputs $\y_i$. 
A model represented by a mapping function $\mathcal{M}$ with parameters $\Theta$, maps an input item $\x$ to the corresponding predicted output $\hat{\y}=\mathcal{M}(\x\vert\Theta)$.

To quantify how strongly the predicted output $\hat{\y}$ deviates from the target output $\y$, a loss function $\mathcal{L}(\hat{\y}, \y)$ is defined. Training the model then comes down to minimizing the expected loss $\bar{\mathcal{L}}=\frac{1}{N}\sum_i \mathcal{L}\big(\mathcal{M}(\x_i| \Theta), \y_i\big)$ over the training set.
In the specific setting of multiclass classification, each input instance corresponds to one out of a fixed set of $M$ output categories. The target vectors $\y$ are typically represented as one-hot vectors: all components are zero, except at index $m$ of the corresponding category. The predicted counterpart $\hat{\y}$ at the model's output is often obtained by applying a so-called softmax output function to intermediate real-valued scores $\mathbf{s}$ 
obtained at the output of the neural network.

The $i$'th component of the softmax is defined as
\[
\text{softmax}(\mathbf{s})_i = \hat{y}_i
= \frac{e^{s_i}}{\sum_j e^{s_j}}
\]
The softmax outputs are well-suited to model a probability distribution (i.e., $0<\hat{y}_i<1$ and $\sum_i\hat{y}_i=1$). 
The standard corresponding loss function is the cross-entropy loss, which quantifies the deviation between the empirical output distribution $\hat{\y}$ (i.e., the softmax outputs) and the ground truth distribution (i.e., the one-hot target vector $\y$)
defined as
\[
\mathcal{L}= - \sum_j y_j \log \hat{y}_j
\]

The most widely used optimization approaches for neural networks are variations of the gradient descent algorithm, in which the parameters $\Theta$ are iteratively updated by taking small steps along the negative gradient of the loss. An estimate $\Theta_n$ at iteration $n$ is updated as $\Theta_{n+1} = \Theta_n - \lambda \nabla_\Theta\ \bar{\mathcal{L}}$, in which the step size is controlled by the learning rate $\lambda$. Typically, training is not performed over the entire dataset per iteration, but instead over a smaller `mini-batch' of instances. This is computationally more efficient and allows for a better exploration of parameter space. 
Importantly, the loss gradient can only be calculated if all components of the neural network are differentiable. 

A deep neural network typically has a layer-wise architecture: the different layers correspond to nested differentiable functions in the overall mapping function $\mathcal{M}$.  
The `forward pass' through the network corresponds to consecutively applying these layer functions to a given input to the network. The intermediate representations obtained by evaluating these layer functions are called hidden states. After a forward pass, the gradient with respect to all parameters can then be calculated by applying the chain rule. This happens during the so-called `backward pass': the gradients are calculated from the output back to the first layer.
As an illustration of how the chain rule is applied, consider the network function $\mathcal{M}(\x| \Theta) = \mathbf{g}\big(\mathbf{f}(\x, \theta_f), \theta_g\big)$, which contains a first layer represented by the vector function $\mathbf{f}$, and a second layer $\mathbf{g}$. For simplicity, say each layer has one trainable parameter, respectively written as $\theta_f$ and $\theta_g$. The derivative with respect to these parameters of a scalar loss function applied to the network output, becomes
\[
\nabla_\Theta\mathcal{L}\big(\mathcal{M}(\x| \Theta)\big) 
= \bigg[\frac{d \mathcal{L}}{d \theta_f},\;\frac{d \mathcal{L}}{d \theta_g}\bigg]\\
= \bigg[\sum_i \frac{\partial \mathcal{L}}{\partial g_i} \sum_j \frac{\partial g_i}{\partial f_j} \frac{\partial f_j}{\partial \theta_f},\; \sum_i \frac{\partial \mathcal{L}}{\partial g_i} \frac{\partial g_i}{\partial \theta_g}
\bigg]
\]
in which the individual derivatives are evaluated based on the considered input $\x$ and current value of the parameters. The entire procedure to calculate the gradients is called the backpropagation algorithm. It requires a forward pass to calculate all intermediate representations up to the value of the loss. After that, in the backward pass, the gradients corresponding to all operations applied during the forward pass, are iteratively calculated, starting at the loss (i.e., with $\partial\mathcal{L}/\partial g_i$ in the example). As such, the gradients with respect to parameters at a given layer can be calculated as soon as the gradients due to all operations further in the network are known, as governed by the chain rule.

To summarize, a single iteration in the optimization happens as follows: 1) A minibatch is sampled from the training data. 2) The output of the neural network is calculated during the forward pass. 3) The loss is calculated based on that output and the target. 4) The gradients for the parameters in the neural network are calculated using backpropagation. 5) The parameters are updated using a gradient-based optimizer. 

The most basic neural networks building block is the so-called fully-connected layer. It consists of a linear transformation with weight matrix $\mathbf{W}$ and bias vector $\mathbf{b}$, followed by applying a component-wise non-linear function, called the activation function. The input into such a layer can be the vector representation $\x$ of the actual input to the model, or the output $\h^{<i>}$ from a previous layer $i$, which is called a hidden representation. Its output is calculated as $\h^{<i+1>} = a(\mathbf{W}\h^{<i>} + \mathbf{b})$, in which typical choices for the activation function $a$ are the Rectified Linear Unit (ReLU) defined as $a(x)=\max(0, x)$ or the hyperbolic tangent $a(x)=\tanh(x)$. In other cases, a sigmoid activation can be used, given by $\sigma(x) = (1+e^{-x})^{-1}$. 
Another important type of neural network layer is the convolutional layer, which convolves the input to pass it to the next layer, by means of a kernel with trainable weights, typically much smaller than the input size. Convolutional layers, followed by a similar activation function, are often used in image recognition models, whereby subsequent layers learn to extract useful features for the given task, from local patterns up to more global and often interpretable patterns. An architecture well-suited for modeling sequences are the so-called recurrent neural networks (RNN). In short, these define a mapping from an input element in the considered sequence into a hidden representation. Every input element is encoded with the same neural network, called the RNN `cell', such that the model can be applied to variable-length sequences. In order to explicitly model the sequential behavior, when encoding a given item in the sequence, the cell takes as input that item, as well as the hidden state obtained while encoding the previous input item. When training with this recurrent setup, the gradient propagates back through the entire sequence. When encoding long sequences, this may lead to very small gradients. An important type of RNN, well-equipped to deal with this so-called vanishing gradient problem, is the Long Short-term Memory (LSTM). The same problem is solved by a similar architecture called the GRU.

More technical details, as well as several other popular types of neural network components, are provided in \cite{Goodfellow2016}.

Deep neural networks can become very expressive, especially when deeper, or with large hidden representations. To avoid overfitting, various regularization approaches have been developed. A widespread technique, also used in some of the presented experiments in this work, is called dropout. For those layers on which dropout is applied, during training a random sample of the layer outputs are set to zero in each iteration, while accordingly compensating the amplitude of the remaining activations. 
At inference time, i.e., when applying the trained model to held-out data, all activations are kept.

As mentioned, many choices are possible in terms of architecture and training: dimensions, types of layers, learning rates, regularization strength, etc. These are so-called hyper-parameters, and are typically `tuned' by evaluating on a validation set, not used explicitly for gradient-based training of the network parameters, and still separate from the final test data.

\section{Introducing DeepProbLog}
\label{sec:deepproblog}
We now recall the basics of probabilistic logic programming using ProbLog (see \citet{deraedt15} for more details), and then introduce our new language DeepProbLog. 
\subsection{ProbLog}
\label{sec:problog}
\begin{definition}[ProbLog program]
A ProbLog program consists of a set of ground probabilistic facts  $\mathcal{F}$ of the form $p\prob f$ where $p$ is a probability and $f$ a ground atom and  a set of rules $\mathcal{R}$.
\end{definition}
For instance, the following ProbLog program models a variant of the well-known alarm Bayesian network \citep{Pearl88:book}:\\
\begin{align*} 
0.1&\prob \mathtt{burglary.}\\
0.5&\prob \mathtt{hears\_alarm(mary)}. \\ 
0.2&\prob \mathtt{earthquake}.\\
0.4&\prob \mathtt{hears\_alarm(john)}. \\ \\
\mathtt{alarm}&\lpif \mathtt{earthquake}.\\ 
\mathtt{alarm}&\lpif \mathtt{burglary}. \\ 
\mathtt {calls(X)}&\lpif \mathtt{alarm, hears\_alarm(X)}.
\end{align*}

Each  probabilistic fact corresponds to an \textit{independent Boolean random variable} that is true with probability $p$ and false with probability $1-p$. Every subset $F \subseteq \mathcal{F}$ defines a possible world $w_F$ = $ F \cup \{ h\theta | \mathcal{R} \cup F \models h\theta$ and $h\theta$ is ground$\}$, that is, the world $w_F$ is the canonical model of the logic program obtained by adding $F$ to the  set of rules $\mathcal{R}$, e.g., 
{\small
\[
w_{\{\mathtt{burglary},\mathtt{hears\_alarm(mary)}\}} = \{\mathtt{burglary},\mathtt{hears\_alarm(mary)}\} \cup \{\mathtt{alarm},\mathtt{calls(mary)}\}
\]
}
To keep the presentation simple, we focus on the case of finitely many ground probabilistic facts, but note that the semantics is also well-defined for the countably infinite case. 
The probability $P(w_F)$ of such a possible world $w_F$ is given by the product of the probabilities of the truth values of the probabilistic facts:
\begin{equation}
P(w_F) = \prod_{f_i \in F}p_i \prod_{f_i \in \mathcal{F} \setminus F}\left(1 - p_i\right)
\end{equation}
For instance, 
\[
P(w_{\{\mathtt{burglary},\mathtt{hears\_alarm(mary)}\}}) = 0.1 \times 0.5\times (1 - 0.2) \times (1-0.4) = 0.024
\]
The probability of a ground fact $q$, also called \emph{success probability of $q$}, is then defined as the sum of the probabilities of all worlds containing $q$, i.e.,
\begin{equation}
P(q) = \sum_{F \subseteq \mathcal{F} : q \in w_F}  P(w_F)
\end{equation}
The probability of a  query is  also equal to the weighted model count (WMC) of the worlds where this query is true.

For ease of modeling, ProbLog supports non-ground probabilistic facts as a shortcut for introducing a set of ground probabilistic facts, as well as annotated disjunctions (ADs), which are expressions of the form
\[
p_1 \prob h_1~;~...~;~p_n \prob h_n~\lpif~b_1, ..., b_m.
\]
where the $p_i$ are probabilities that sum to at most one, the $h_i$ are atoms, and the $b_j$ are literals.
The meaning of an AD is that whenever all $b_i$ hold, the AD causes one of the $h_j$ to be true, or none of them with probability $1-\sum p_i$. Note that several of the $h_i$ may be true at the same time if they also appear as heads of other rules or ADs.
This is convenient to model choices between different categorical variables, e.g. different severities of the earthquake:
\[
0.4\prob \mathtt{earthquake(none)}~;~0.4\prob \mathtt{earthquake(mild)}~;~0.2\prob \mathtt{earthquake(severe)}.
\]
or without explicitly representing the event of no earthquake:
\[
0.4\prob \mathtt{earthquake(mild)}~;~0.2\prob \mathtt{earthquake(severe)}.
\]
In which neither \texttt{earthquake(mild)} nor \texttt{earthquake(severe)} will be true with probability $0.4$.
Annotated disjunctions do not change the expressivity of ProbLog, as they can alternatively be modeled through independent facts and logical rules; we refer to \citet{deraedt15} for technical details.

\subsection{DeepProbLog}

In ProbLog, the probabilities of all random choices are explicitly specified as part of probabilistic facts or annotated disjunctions. DeepProbLog extends ProbLog to basic random choices whose probabilities are specified through external functions implemented as neural networks. 

\begin{definition}[Neural annotated disjunction]
A \emph{neural AD} is an expression of the form 
\[
nn(m_r, \vec{I},O,\vec{d}) \prob r(\vec{I},O).
\]
where $nn$ is a reserved functor, $m_r$ uniquely identifies a neural network model with $k$ inputs and $n$ outputs (i.e., its architecture as well as its trainable parameters) defining a probability distribution over $n$ classes, $\vec{I}=I_1,...,I_k$ is a sequence of input variables, $O$ is the output variable, $\vec{d}=d_1,...,d_n$ is a sequence of ground terms (the output domain of this neural network) and $r$ is a predicate.\\\\
A \emph{ground neural AD} is an expression of the form
\[
nn(m_r,\vec{i},d_1) \prob r(\vec{i},d_1)~;~...~;~ nn(m_r,\vec{i},d_n) \prob r(\vec{i},d_n).
\]
where $\vec{i}=i_1,...,i_k$ is a sequence of ground terms (the input to the neural network) and $d_1,...,d_n$ are ground terms (the output domain of this neural network).
\end{definition}

The $nn(m_r,\vec{i},d_j)$ term in the definition can be considered a function that returns the probability of class $d_j$ when evaluating the network $m_r$ on input $\vec{i}$.
As such, a ground nAD can be instantiated into a normal AD by evaluating the neural network and replacing the functor with the calculated probability. 
For instance, in the MNIST addition example, we would specify the nAD
\[
\mathtt{nn(m\_{digit},[X],Y,[0,\ldots,9]) \prob digit(X,Y).}
\]
where \texttt{m\_digit} is a network that classifies MNIST digits. Grounding this on an input image $\digit{mnist_3}$ would result in a ground nAD:
\[
\mathtt{nn(m\_{digit},[\digit{mnist_3}],0) \prob digit(\digit{mnist_3},0)~;~\ldots~;~nn(m\_{digit},[\digit{mnist_3}],9) \prob digit(\digit{mnist_3},9).}
\]
Evaluating this would result in a ground AD:
\[
\mathtt{p_0 \prob digit(\digit{mnist_3},0)~;~\ldots~;p_9 \prob digit(\digit{mnist_3},9).}
\]
Where $[p_0, \ldots, p_9]$ is the output vector of the \texttt{m\_digit} network when evaluated on $\digit{mnist_3}$.\\
The neural network could take any shape, e.g., a convolutional network for image encoding, a recurrent network for sequence encoding, etc. However, its output layer, which feeds the corresponding neural predicate, needs to be normalized.\\

We consider an output domain size of two as a special case. Instead of the neural network having two probabilities at the output that sum to one, we can simplify this to a single probability, with the second one the complement of that probability. This difference coincides with the difference between a softmax and single-neuron sigmoid layer in a neural network. We call such an expression a neural fact.
\begin{definition}[Neural fact]
A neural fact is an expression of the form
\[
nn(m_r,\vec{I}) \prob r(\vec{I}).
\]
where $nn$ is a reserved functor, $m_r$ uniquely identifies a neural network model defining a probability distribution over $n$ classes, $\vec{I}=I_1,...,I_k$ is a sequence of input variables and $r$ is a predicate.\\\\
A ground neural fact is an expression of the form
\[
nn(m_r,\vec{i}) \prob r(\vec{i}).
\]
 where $\vec{i}=i_1,...,i_k$ is a sequence of ground terms (the input to the neural network).
\end{definition}

To exemplify, we use a neural network that gives a measure of the similarity between two input images. We can encode this with the following neural fact:
\[
\mathtt{nn(m, [X,Y]) \prob similar(X,Y).}
\]
Grounding this on the input $\digit{mnist_3}$ and $\digit{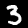}$ would result in the follow ground neural fact:
\[
\mathtt{nn(m, [\digit{mnist_3},\digit{mnist_3b}]) \prob similar(\digit{mnist_3},\digit{mnist_3b}).}
\]
Evaluating this would result in a ground probabilistic fact:
\[
\mathtt{p \prob similar(\digit{mnist_3},\digit{mnist_3b}).}
\]
Where $p$ is the output of the $m$ network when evaluated on $\digit{mnist_3}$ and $\digit{mnist_3b}$.
\begin{definition}[DeepProbLog Program]
A DeepProbLog program consists of a set of ground probabilistic facts  $\mathcal{F}$, a set of ground neural ADs and ground neural facts $\mathcal{N}$, and a set of rules $\mathcal{R}$.
\end{definition}
The semantics of a DeepProbLog program is given by the semantics of the ProbLog program obtained by replacing each nAD with the AD obtained by instantiating the probabilities as mentioned above. While the semantics is defined with respect to ground neural ADs and facts, as in ProbLog, we write non-ground such expressions if the intended grounding is clear from context.

\section{Inference}
This section explains how a DeepProbLog model is used for a given query at prediction time.
First, we provide more detail on ProbLog inference \citep{Fierens}. Next, we describe how ProbLog inference is adapted in DeepProbLog. 
\subsection{ProbLog Inference}\label{sec:inference}
ProbLog inference proceeds in four steps. 
The first step is the grounding step, in which the logic program is grounded with respect to the query. This step uses backward reasoning to determine which ground rules are relevant to derive the truth value of the query, and may perform additional logical simplifications that do not affect the query's probability.\\\\
The second step rewrites the ground logic program into a formula in propositional logic that defines the truth value of the query in terms of the truth values of probabilistic facts. We can calculate the query success probability by performing \emph{weighted model counting} (WMC) on this logic formula (cfr. \citet{Fierens}). However, performing WMC on this logical formula directly is not efficient.\\\\
The third step is knowledge compilation \citep{DarwicheMarquis}. During this step, the logic formula is transformed into a form that allows for efficient weighted model counting.
The current ProbLog system uses Sentential Decision Diagrams (SDDs, \citet{darwiche2011sdd}), the most succinct suitable representation available today.
SDDs, being a subset of d-DNNFs allow for polytime model counting (\cite{DarwicheMarquis}). However, they also support polytime conjunction, disjunction and negation while being more succinct than OBDDs (\citet{darwiche2011sdd}).\\

The fourth and final step transforms the SDD into an arithmetic circuit (AC). This is done by putting the probabilities of the probabilistic facts or their negations on the leaves, replacing the OR nodes with addition and the AND nodes by multiplication. The WMC is then calculated with an evaluation of the AC.
\begin{figure}[p]
    \centering
    \begin{subfigure}[b]{0.45\linewidth}
        \centering
        \begin{minted}[frame=lines, fontsize=\small]{Prolog}
0.2::earthquake.
0.1::burglary.
0.5::hears_alarm(mary).
0.4::hears_alarm(john).
alarm :- earthquake.
alarm :- burglary.
calls(X):-alarm,hears_alarm(X).
        \end{minted}
        \caption{The ProbLog program.}
        \label{fig:program_nonground}
    \end{subfigure}
    \begin{subfigure}[b]{0.45\linewidth}
        \centering
        \begin{minted}[frame=lines, fontsize=\small]{Prolog}
0.2::earthquake.
0.1::burglary.
0.5::hears_alarm(mary).

alarm :- earthquake.
alarm :- burglary.
calls(mary):-alarm,hears_alarm(mary).
        \end{minted}
        \caption{The relevant ground program.}
        \label{fig:program_grounded}
    \end{subfigure}
    
    \vspace{15pt}
    \begin{subfigure}[b]{\linewidth}
    \centering
    \begin{tabular}{l|r}
    Models of {\footnotesize $\mathtt{calls(mary)} \leftrightarrow \mathtt{hears\_alarm(mary)} \wedge (\mathtt{burglary} \vee \mathtt{earthquake})$} & w \\ \hline
    \{\} & 0.36 \\
    \{\texttt{hears\_alarm(mary)}\} & 0.36 \\
    \{\texttt{earthquake}\} & 0.09 \\
    \{\texttt{earthquake, hears\_alarm(mary),calls(mary)}\} & \textbf{0.09} \\
    \{\texttt{burglar}y\} & 0.04 \\
    \{\texttt{burglary, hears\_alarm(mary),calls(mary)}\} & \textbf{0.04} \\
    \{\texttt{burglary, earthquake}\} & 0.01 \\
    \{\texttt{burglary, earthquake, hears\_alarm(mary),calls(mary)}\} & \textbf{0.01} \\ \hline \hline
     \multicolumn{1}{r}{$\sum_{\mathtt{calls(mary)} \in \text{model}}$}& \textbf{0.14}
    \end{tabular}
    \caption{The weighted count of the models where \texttt{calls(mary)} is true.}
    \label{fig:wmc}
    \end{subfigure}
    
    \vspace{15pt}
    \begin{subfigure}[b]{\linewidth}
        \centering
        \includegraphics[width=0.8\linewidth]{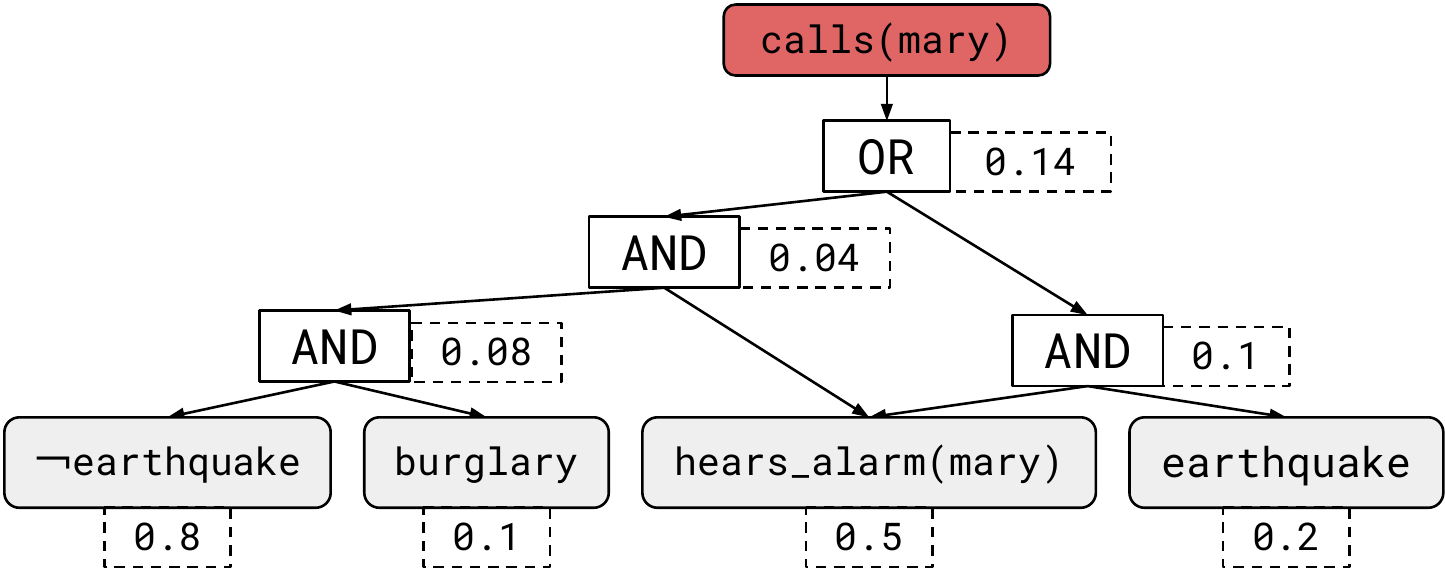}
        \caption{The AC for query \texttt{calls(mary)}.}
        \label{fig:program_compiled}
    \end{subfigure}
    \caption{Inference in ProbLog using query \texttt{calls(mary)} and the program in (a). (Example~\ref{ex:inference})}
    \label{fig:inference}
\end{figure}

\begin{example} \label{ex:inference}
In Figure \ref{fig:inference}, we apply the four steps of ProbLog inference on the earthquake example  
with query \texttt{calls(mary)}.\\
In the first step, the non-ground program (Figure~\ref{fig:program_nonground}) is grounded with respect to the query \texttt{calls(mary)}. The result is shown in Figure~\ref{fig:program_grounded}: the irrelevant fact \texttt{hears\_alarm(john)} is omitted and the variable \texttt{X} in the \texttt{calls} rule is substituted  with the constant \texttt{mary}.
The resulting formula in the second step is $$\mathtt{calls(mary)} \leftrightarrow \mathtt{hears\_alarm(mary)} \wedge (\mathtt{burglary} \vee \mathtt{earthquake})$$
The WMC of this formula is shown in Figure~\ref{fig:wmc}. However, it is not calculated by enumeration as shown here, but an AC is used instead. The AC derived in step four is shown in Figure~\ref{fig:program_compiled}, where rounded grey rectangles depict variables corresponding to probabilistic facts, and the rounded red rectangle denotes the query atom defined by the formula.  The white rectangles correspond to logical operators applied to their children. The intermediate results are shown in black next to the nodes in Figure~\ref{fig:program_compiled}.
\end{example}

\subsection{DeepProbLog Inference}
\label{sec:dplinference}
The only change required for DeepProbLog inference is that we need to instantiate the ground nADs and neural facts into the corresponding ground ADs and ground facts. This is done in a separate step after grounding, where the parameters for the regular AD are determined by making a forward pass on the relevant neural network with the ground input.
\begin{example}\label{ex:addition}
We illustrate this by evaluating the MNIST addition example (Figure~\ref{fig:dpl_nonground}).
The DeepProbLog program requires two lines: the first line defining the neural predicate, and the second line defining the addition. 
We evaluate it on the query $\mathtt{addition(\digit{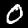}, \digit{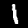}, 1)}$. In the first step, the DeepProbLog program is grounded into a ground DeepProbLog Program (Figure~\ref{fig:dpl_ground}).
Note that the nADs are now all ground. As ProbLog only grounds the relevant part of the program, i.e. the part that can be used to prove the query, only the digits $0$ and $1$ are retained as the larger digits cannot sum to $1$. The next step is the only difference between ProbLog and DeepProbLog inference: instantiating the ground nADs into regular ground ADs, which could, for instance, produce an AD as shown in Figure~\ref{fig:dpl_prob}.
The probabilities in the instantiated ADs do not sum to one, as the  irrelevant terms ($\mathtt{digit(\digit{0},2)}$, ...,$\mathtt{digit(\digit{0},9)}$ and $\mathtt{digit(\digit{1},2)}$, ..., $\mathtt{digit(\digit{1},9)}$) have been dropped in the grounding process, although the neural network still assigns probability mass to them.
Inference then proceeds identically to that of ProbLog: the ground program is rewritten into a logical formula, this formula is compiled and transformed into an AC. Finally, this AC is evaluated to calculate the query probability.
\end{example}
\begin{figure}[t!]
\centering
    \begin{subfigure}[b]{0.9\linewidth}
        \centering
        \begin{prolog}
nn(m_digit, [X], Y, [0...9]) :: digit(X,Y).
addition(X,Y,Z) :- digit(X,N1), digit(Y,N2), Z is N1+N2.
        \end{prolog}
        \caption{The DeepProbLog program.}
        \label{fig:dpl_nonground}
    \end{subfigure}
    
    \vspace{10pt}
    \begin{subfigure}[b]{\linewidth}
        \centering
        \begin{prolog}
nn(m_digit,[d0],0)::digit(d0,0);nn(m_digit,[d0], 1)::digit(d0,1).
nn(m_digit,[d1],0)::digit(d1,0);nn(m_digit,[d1], 1)::digit(d1,1).
addition(d0,d1,1) :- digit(d0,0), digit(d1,1).
addition(d0,d1,1) :- digit(d0,1), digit(d1,0).
        \end{prolog}
        \caption{The ground DeepProbLog program.}
        \label{fig:dpl_ground}
    \end{subfigure}
    
    \vspace{10pt}
    \begin{subfigure}[b]{0.75\linewidth}
        \centering
        \begin{prolog}
0.8 :: digit(d0,0); 0.1 :: digit(d0,1).
0.2 :: digit(d1,0); 0.6 :: digit(d1,1).
addition(d0,d1,1) :- digit(d0,0), digit(d1,1).
addition(d0,d1,1) :- digit(d0,1), digit(d1,0).
        \end{prolog}
        \caption{The ground ProbLog program.}
        \label{fig:dpl_prob}
    \end{subfigure}
    \caption{Inference in DeepProbLog (Example~\ref{ex:addition})}
    \label{fig:dpl_inference}
\end{figure}

\section{Learning in DeepProbLog}
\label{sec:Learning}
We now introduce our approach to learn the parameters in DeepProbLog programs. The parameters include the learnable parameters of the neural network (which we will call neural parameters from now on) and the learnable parameters in the logic program (which we will refer to as probabilistic parameters).
We use the \emph{learning from entailment} setting~\citep{lfe}
\begin{definition}{\emph{Learning from entailment}}\label{def:entailment}
Given a DeepProbLog program with parameters $\Theta$, a set $\mathcal{Q}$ of pairs $(q,p)$ with $q$ a query and  $p$ its desired success probability, and a loss function $\mathcal{L}$, compute:
\[
\argmin_{\Theta} \frac{1}{|\mathcal{Q}|}\sum_{(q,p)\in \mathcal{Q}} \mathcal{L}(P(q|\Theta),p)
\]
\end{definition}

In most of the experiments, unless mentioned otherwise, we only use positive examples for training (i.e., with desired success probability $p=1$). The model then needs to adjust the weights to maximize query probabilities $P_{\Theta}(q)$ for all training examples. This can be expressed by minimizing the average negative log likelihood of the query, whereby 
Definition~\ref{def:entailment} reduces to:
\[
\argmin_{\Theta} \frac{1}{|\mathcal{Q}|}\sum_{(q,p)\in \mathcal{Q}} -\log P_{\Theta}(q)
\]
The presented method however works for other choices in the loss function. For example, in experiment \textbf{T9} (Section~\ref{subsec:probprogr}) the mean squared error (MSE) is used.

\subsection{Gradient descent in ProbLog}
In contrast to the earlier approach for ProbLog parameter learning in this setting by  \citet{gutmann2008parameter}, we use gradient descent rather than EM. This allows for seamless integration with neural network training.
The key insight here is that we can use the same AC that ProbLog uses for inference for gradient computations as well. 
We rely on the automatic differentiation capabilities already available in ProbLog to derive these gradients.
More specifically, to compute the gradient with respect to the probabilistic logic program part, we rely on Algebraic ProbLog (aProbLog \citep{kimmig2011algebraic}), a generalization of the ProbLog language and inference to arbitrary commutative semirings, including the gradient semiring \citep{eisner2002parameter}. 
In the following, we provide the necessary background on aProbLog, 
discuss how to use it 
to compute gradients with respect to ProbLog parameters and extend the approach to DeepProbLog. 
\paragraph{aProbLog and the gradient semiring}\label{sec:gradientsemiring}
ProbLog annotates each probabilistic fact $f$ with the probability $P$ that $f$ is true, which implicitly also defines the probability $1-P$ that 
its negation $\neg f$ is true. 
It then uses the probability semiring with regular addition and multiplication as operators to compute the probability of a query on the AC constructed for this query, cf.~Figure~\ref{fig:program_compiled}.
The probability semiring is defined as follows:
\begin{align}
a \oplus b &= a + b  \\
a \otimes b &= ab\\
e^\oplus &= 0\\
e^\otimes &=1 
\end{align}
And the accompanying labeling function as:
\begin{align}
L(f) &=  p &&\text{for}~p\prob f\\
L(\neg f) &= 1-p &&\text{with}~L(f) = p
\end{align}
This idea is generalized in aProbLog to compute such values based on arbitrary commutative semirings. 
Instead of probability labels on facts, aProbLog uses a labeling function that explicitly associates values from the chosen semiring with both facts and their negations, and combines these using semiring addition~$\oplus$ and multiplication~$\otimes$ on the AC. 
We  use  the gradient semiring, whose elements are tuples $(p,\frac{\partial p}{\partial \theta})$, where $p$ is a probability (as in ProbLog), and $\frac{\partial p}{\partial \theta}$ is the  partial derivative of that probability with respect to a parameter $\theta$, that is, the probability $p_i$ of a probabilistic fact  with learnable probability, written as $t(p_i)\prob f_i$. 
This is easily extended  to a vector of parameters $\vec{\theta} = [\theta_1,\ldots,\theta_N]^T$, the concatenation of all $N$ probabilistic parameters in the ground program, as it is easier and faster to process all gradients in one vector.
Semiring addition~$\oplus$, multiplication~$\otimes$ and the neutral elements with respect to these operations are defined as follows:

\begin{align}
(a_1,\vec{a_2}) \oplus (b_1,\vec{b_2}) &= (a_1+b_1, \vec{a_2} +  \vec{b_2})  \\
(a_1,\vec{a_2}) \otimes (b_1,\vec{b_2}) &= (a_1b_1, b_1\vec{a_2} +  a_1\vec{b_2})\\
e^\oplus &= (0,\vec{0})\\
e^\otimes &= (1,\vec{0})
\end{align}

Note that the first element of the tuple mimics ProbLog's probability computation, whereas the second simply computes gradients of these probabilities using derivative rules.
\paragraph{Gradient descent with aProbLog} To use the gradient semiring for gradient descent parameter learning in ProbLog, we first transform the ProbLog program into an aProbLog program by extending the label of each probabilistic fact $p\prob f$ to include the probability $p$ as well as the gradient vector of $p$ with respect to the probabilities of all probabilistic facts and ADs in the program, i.e.,  
\begin{align}
L(f) &=  (p,\vec{0}) &&\text{for}~p\prob f~\text{with fixed}~p\\
L(f_i) &=  (p_{i}, \mathbf{e}_i) &&\text{for}~t(p_i)\prob f_i~\text{with learnable}~p_i\\
L(\neg f) &= (1-p, - \nabla p) &&\text{with}~L(f) = (p, \nabla p)
\end{align}
where the vector  $\mathbf{e}_i$ has a $1$ in the $i$th position and $0$ in all others. 
For fixed probabilities, the gradient does not depend on any parameters and thus is 0. Note that after each update step, the probabilistic parameters are clipped to the $[0,1]$ range, and the parameters of an AD are re-normalized to ensure that they sum to one.
For the other cases, we use the semiring labels as introduced above.
\begin{figure}[t!]
\centering
\includegraphics[width=0.7\linewidth]{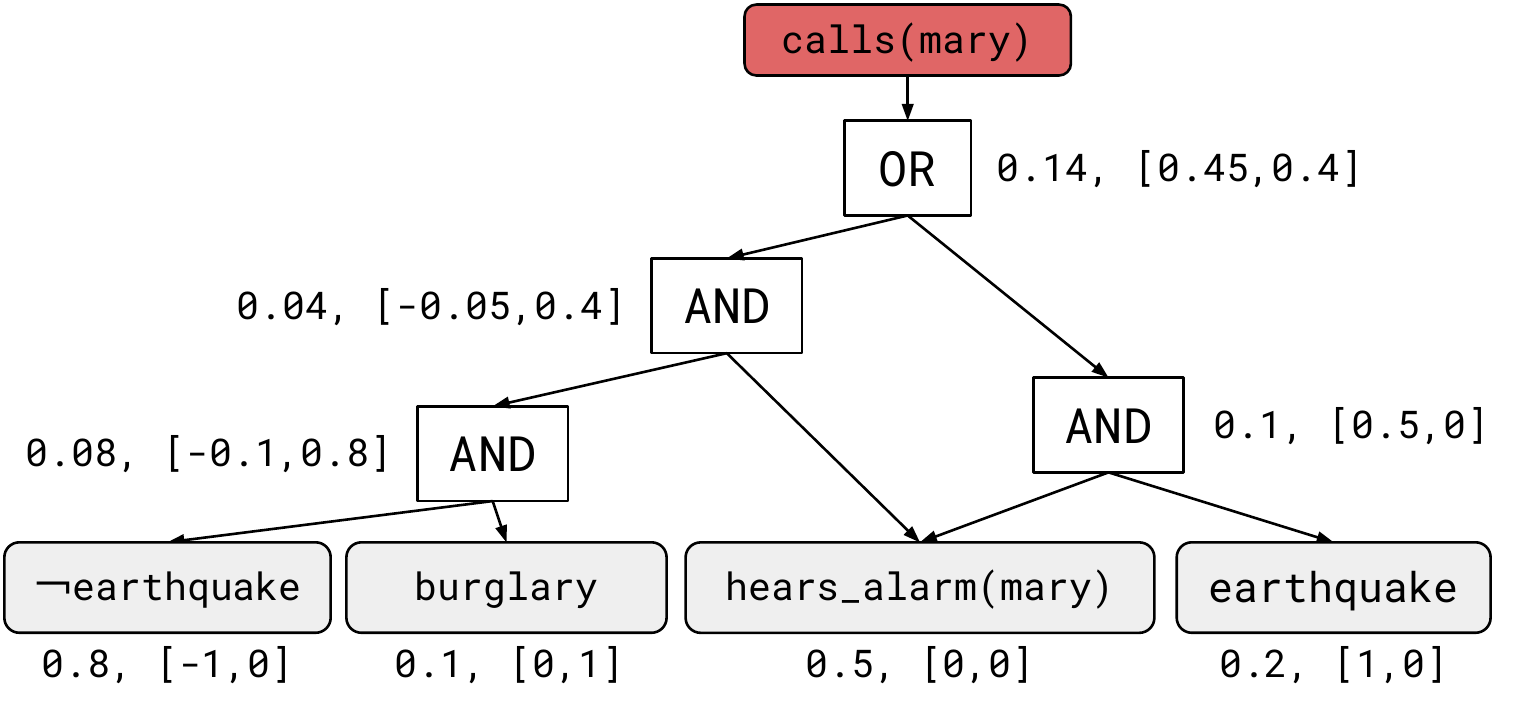}
\caption{The AC evaluated using the gradient semiring. (Example~\ref{ex:learning})}
\label{fig:sdd_grad}
\end{figure}
\begin{example}\label{ex:learning}
Assume we want to learn the probabilities of \verb|earthquake| and \verb|burglary| in the example of Figure~\ref{fig:inference}, while keeping those of the other facts fixed.  Figure~\ref{fig:sdd_grad} shows  the evaluation of the same AC as in Figure~\ref{fig:program_compiled}, but with the gradient semiring. The nodes in the AC now also contain the gradient (the second element of the tuple). The result on the top node shows that the partial derivative of the  query is $0.45$ and $0.4$ w.r.t. the earthquake and burglary parameters respectively.
\end{example}
\subsection{Gradient descent for DeepProbLog}\label{subsec:GD_DPL}
Just as the only difference between inference in ProbLog and DeepProbLog is the evaluation of the nADs, the only difference between gradient descent in ProbLog and DeepProbLog is optimizing the neural parameters alongside the probabilistic parameters.
As mentioned in the previous section, the probabilistic parameters $p_i$ in the logic program can be optimized by using the gradient semiring, which allows us to calculate $\partial P(q)/\partial p_i$. This gradient is then used to perform the update by using gradient descent.
Note that  since the outputs of the neural network are used as probabilities in the logic program and can be learned, we can view them as a kind of \emph{abstract} parameters. However, although we can derive a gradient for these abstract parameters, we cannot optimize them directly, as the logic is unaware of the neural parameters that determine the value of these abstract parameters. 
Recall from Equation (1) that the gradient of the internal (neural) parameters in standard supervised learning can be derived using the chain rule in backpropagation. Below, we show how we can derive the gradient for these neural parameters of the loss applied to $P(q)$ (Definition~\ref{def:entailment}), rather than a loss function defined directly on the output of the neural network.

Specifically, consider the case of a single neural annotated disjunction, with probabilities $\hat{p}_i$ (i.e., the aforementioned abstract parameters), calculated by evaluating a neural network with softmax output. The predicted probability that the query holds true, based on the current values of the neural and probabilistic parameters, is written $P(q)$. 
While training, true examples should yield a predicted query probability close to the expected query probability, which is expressed by means of a loss function $\mathcal{L}$ as introduced in Definition~\ref{def:entailment}. 

Application of the chain rule leads to
\[
\frac{d \mathcal{L}}{d \theta_k} = 
\frac{\partial \mathcal{L}}{\partial P(q)}\;
\sum_i\frac{\partial P(q)}{\partial \hat{p}_i}
\frac{\partial \hat{p}_i}{\partial \theta_k}
\]
where the derivative of the loss with respect to any trainable parameter $\theta_k$ in the neural network is decomposed into the partial derivative of the loss with respect to the predicted output $P(q)$, the latter's derivative $\partial P(q)/\partial \hat{p}_i$ with respect to each component of the annotated disjunction as obtained with the gradient semiring, and finally $\partial \hat{p}_i/\partial \theta_k$, the derivative of the neural network's output components with respect to the considered parameter. The latter is obtained by the standard application of the chain rule in the neural network. The backpropagation procedure in the neural network can thus be started by providing $\partial P(q)/\partial \hat{p}_i$, to systematically obtain the loss gradients for all neural parameters.\\
Extending this approach to the situation of multiple neural predicates is straightforward. If the same neural network is used for different neural predicates (e.g. in Example~\ref{ex:addition}), the final derivative is obtained by summing over the contributions of each neural predicate.

Then, standard gradient-based optimizers (e.g. SGD, Adam, ...) are used to update the parameters of the network. During gradient computation with aProbLog, the probabilities of neural ADs are kept constant. Furthermore, updates on neural ADs come from the neural network part of the model, where the use of a softmax output layer ensures  a normalized distribution, hence not requiring the additional normalization as for non-neural ADs.

To extend the gradient semiring to DeepProbLog programs, we define it  for nADs and neural facts.
The label for the nAD is defined as:
\begin{align}
    L(f_i) = (\hat{p}_j, \mathbf{e}_j)&&\text{for}~\ldots~;~nn(m,\vec{i},d_j) \prob r(\vec{i},d_j) ~;~\ldots~\text{a ground nAD}
\end{align}

Where $d_j$ is the j-th domain element, $\hat{p}_j,$ is the j-th element of the output of the neural network $m$ evaluated on input $\vec{i}$.
The label for a neural fact is defined as:
\begin{align}
    L(f_i) = (\hat{p}, \mathbf{e}_j)&&\text{for}~nn(m,\vec{i}) \prob r(\vec{i})~\text{a ground neural fact}
\end{align}
where $\hat{p}$ is the output of the neural network $m$ evaluated on input $\vec{i}$.
Since the first element of the tuple for nADs and neural facts is the evaluation of the neural networks as in Section~\ref{sec:dplinference}, this change remains semantically equivalent.

\begin{figure}[p]
    \centering
    \begin{subfigure}[b]{\linewidth}
        \centering
        \begin{minipage}{\linewidth}
        \small
        \begin{minted}[frame=lines]{Prolog}
nn(classifier, [X], Y, [0 .. 9]) :: digit(X,Y).
t(0.2) :: noisy.

1/19 :: uniform(X,Y,0) ; ... ; 1/19 :: uniform(X,Y,18).

addition(X,Y,Z) :- noisy, uniform(X,Y,Z).
addition(X,Y,Z) :- \+noisy, digit(X,N1), digit(Y,N2), Z is N1+N2.
        \end{minted}
        \end{minipage}
        \caption{The DeepProbLog program.}
        \label{fig:dproblog_program}
    \end{subfigure}
    
    \begin{subfigure}[b]{\linewidth}
        \begin{minipage}{\linewidth}
        \small
        \begin{minted}[frame=lines]{Prolog}
nn(classifier,[a],0) :: digit(a,0); nn(classifier,[a],1) :: digit(a,1).
nn(classifier,[b],0) :: digit(b,0); nn(classifier,[b],1) :: digit(b,1).
t(0.2)::noisy.

1/19::uniform(a,b,1).
addition(a,b,1) :- noisy, uniform(a,b,1).

addition(a,b,1) :- \+noisy, digit(a,0), digit(b,1).
addition(a,b,1) :- \+noisy, digit(a,1), digit(b,0).
        \end{minted}
        \end{minipage}
        \caption{The ground DeepProbLog program.}
        \label{fig:dproblog_ground}
    \end{subfigure}

    \begin{subfigure}[b]{0.9\linewidth}
        \includegraphics[width=\linewidth]{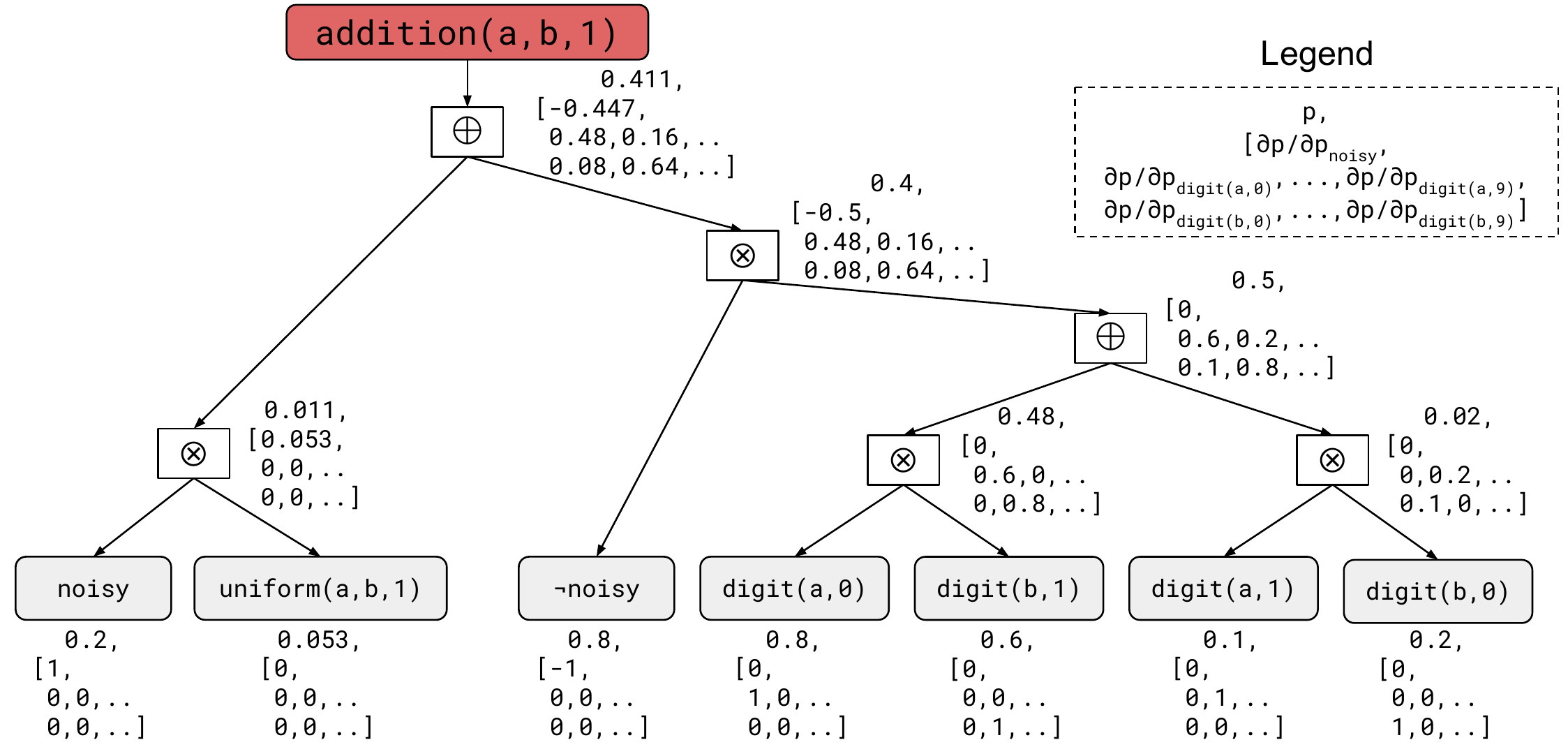}
        \caption{The AC for query \texttt{addition(a,b,1)}.}
        \label{fig:dproblog_compiled}
    \end{subfigure}
    \caption{Parameter learning in DeepProbLog. (Example~\ref{ex:dpl_learning})}
    \label{fig:dpl_learning}
\end{figure}

\begin{figure}
    \centering
    \includegraphics[width=\linewidth]{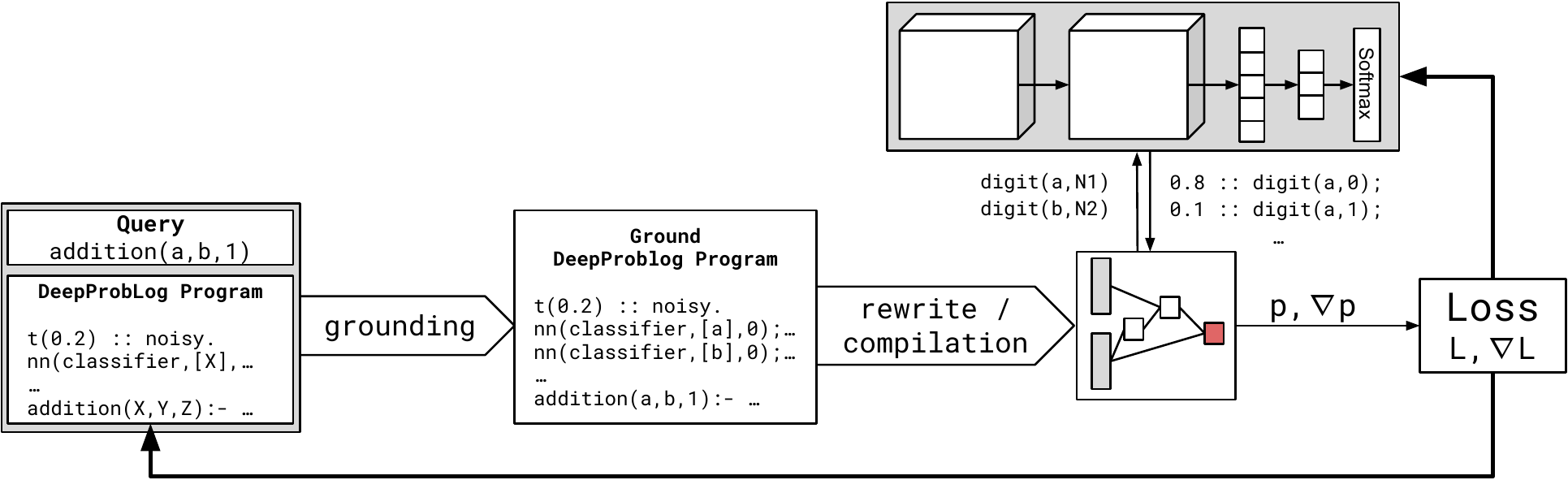}
    \caption{The learning pipeline.}
    \label{fig:pipeline}
\end{figure}

\begin{example}\label{ex:dpl_learning}
To demonstrate the learning pipeline (Figure~\ref{fig:pipeline}), we will apply it on the MNIST addition example show in Section~\ref{sec:dplinference} with a small extension: some of the labels have been  corrupted and  are picked randomly from a uniform distribution over $[0,18]$. The goal is to also learn the fraction of noisy examples. The DeepProbLog program is given in Figure~\ref{fig:dproblog_program}. Grounding on the query $\mathtt{addition(a,b,1)}$ results in the ground DeepProbLog program shown in Figure~\ref{fig:dproblog_ground}. The arithmetic circuit corresponding to the ground program is shown in Figure~\ref{fig:dproblog_compiled}. As can be seen, the neural networks already have a confident prediction for both images (being 0 and 1 respectively). 
The top right shows how the different partial derivatives that are calculated: one w.r.t. to the noisy parameter, ten for the evaluation of the neural network on input a and ten for the evaluation on input b.
\end{example}


\section{Experimental Evaluation}
\begin{figure}[t]
    \centering
    \begin{subfigure}[b]{0.32\linewidth}
        \centering
        \includegraphics[width=\linewidth]{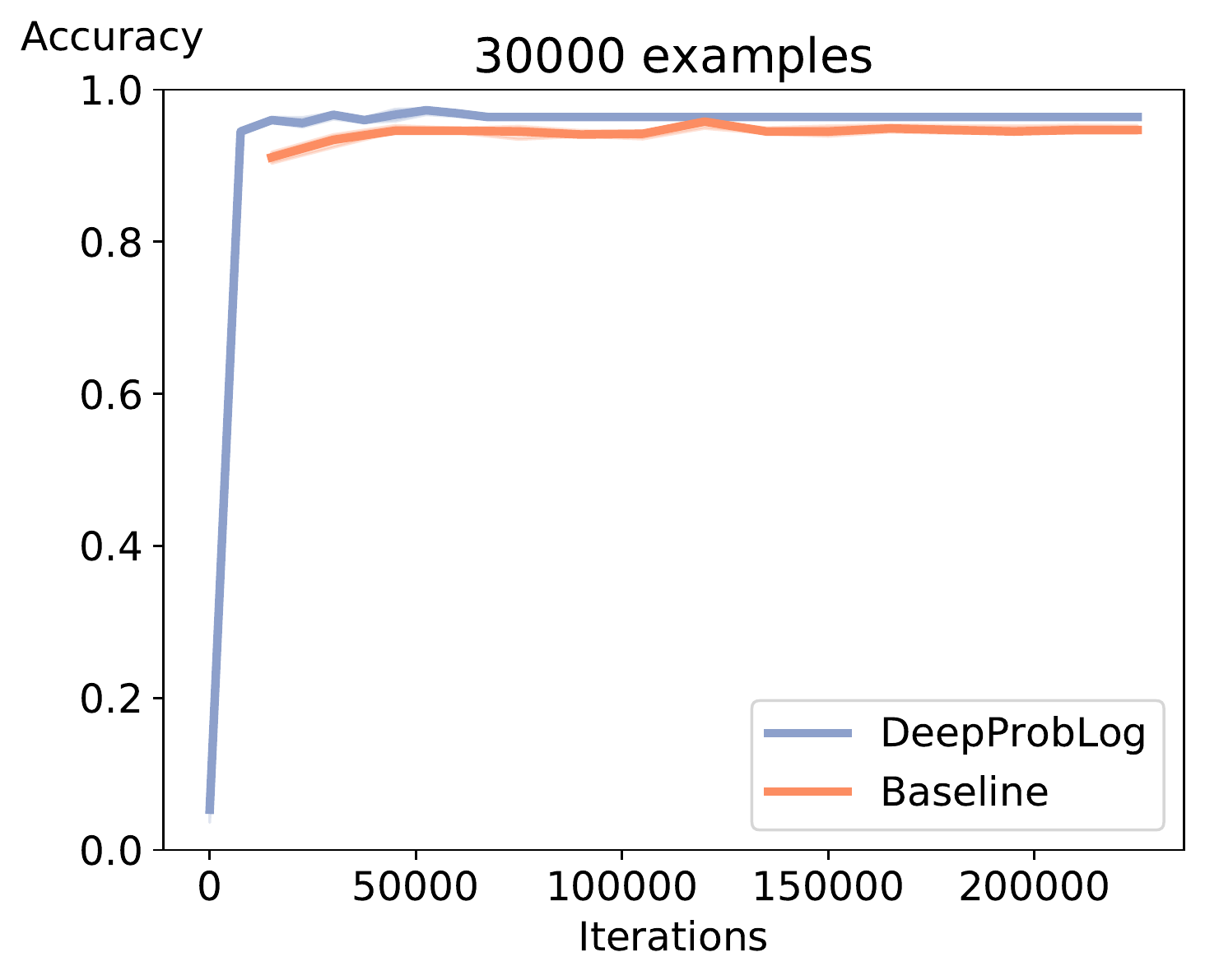}
        \caption{30 000 examples}
        \label{fig:mnist1}
    \end{subfigure}
    \begin{subfigure}[b]{0.32\linewidth}
        \centering
        \includegraphics[width=\linewidth]{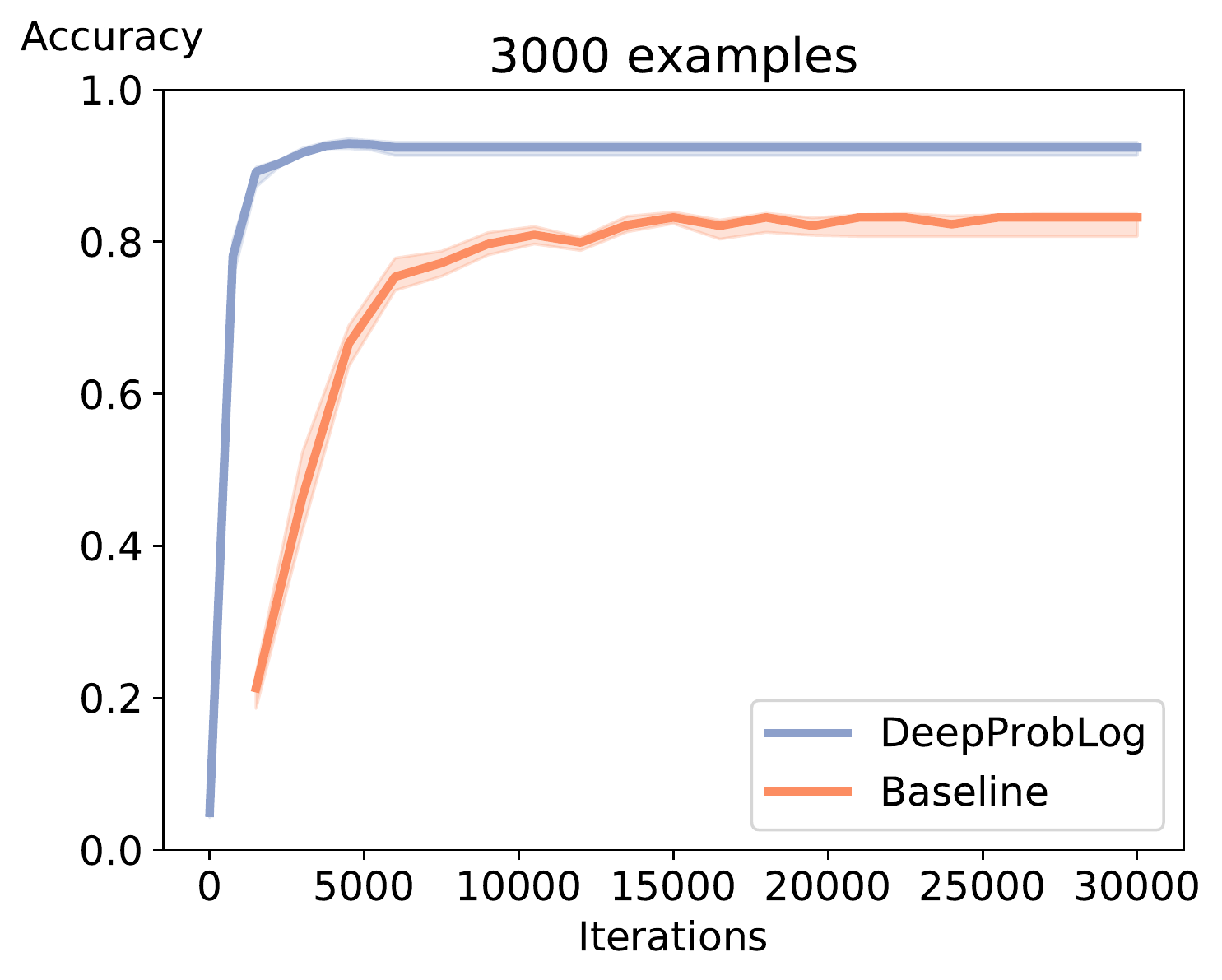}
        \caption{3 000 examples}
        \label{fig:mnist2}
    \end{subfigure}
        \begin{subfigure}[b]{0.32\linewidth}
        \centering
        \includegraphics[width=\linewidth]{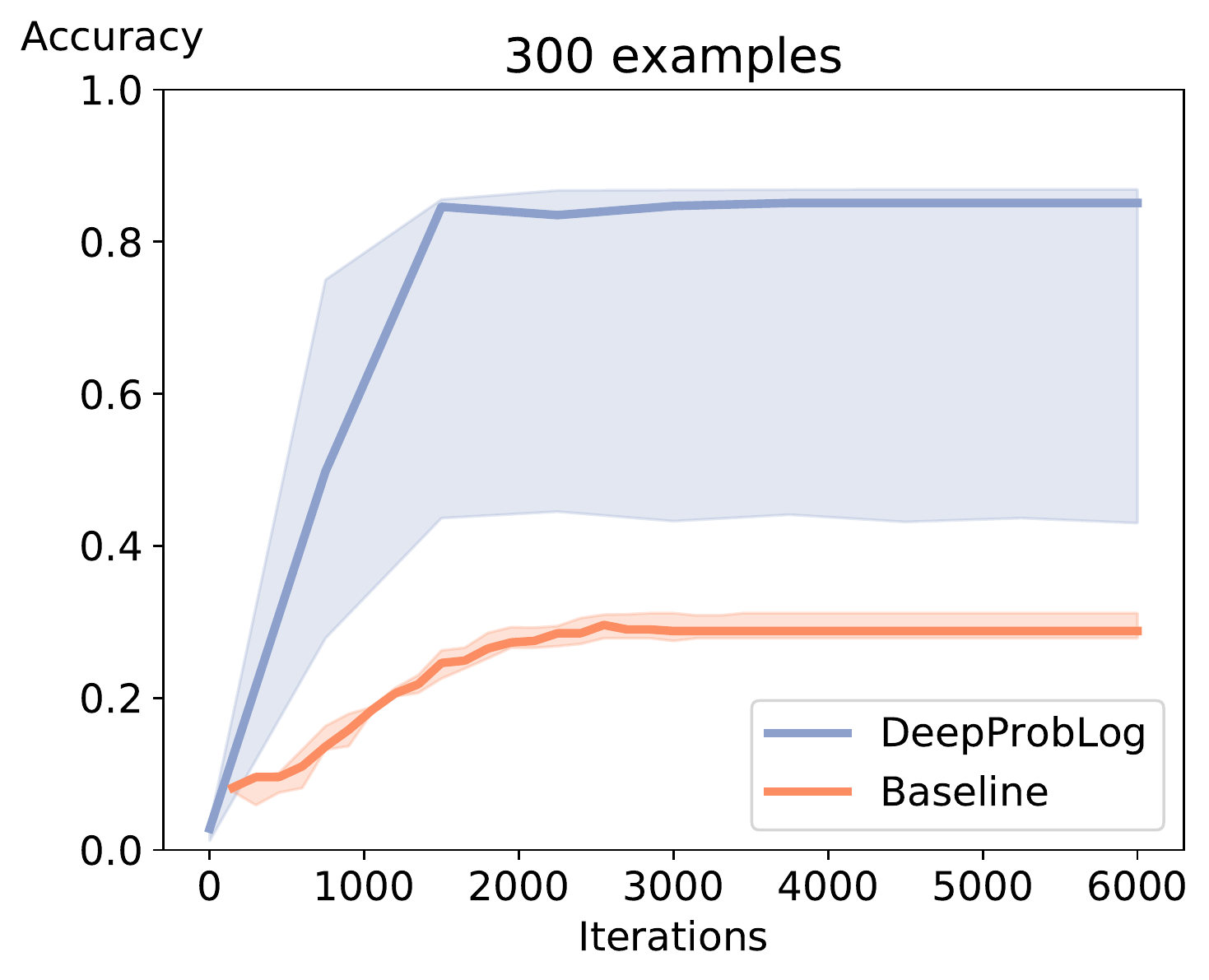}
        \caption{300 examples}
        \label{fig:mnist3}
    \end{subfigure}
    \caption{MNIST Single-Digit Addition (\textbf{T1}). The graphs show the accuracy on the validation set during training for different training set sizes.}
    \label{fig:mnist_single_addition}
\end{figure}

\begin{table}[t]
    \centering
\begin{tabular}{@{}lrrr@{}}
\toprule
            & \multicolumn{3}{c}{Number of training examples}         \\ \cmidrule(l){2-4} 
Model       & 30 000              & 3 000            & 300            \\ \midrule
Baseline    & $93.46 \pm 0.49$ & $78.32 \pm 2.14$ & $23.64 \pm 1.75$ \\
DeepProbLog & $97.20 \pm 0.45$ & $92.18 \pm 1.57$ & $67.19 \pm 25.05$ \\ \bottomrule
\end{tabular}
    \caption{The accuracy on the test set for \textbf{T1}.}
    \label{tab:mnist_single_addition}
\end{table}

\label{sec:experiments}
We perform three sets of experiments to 
demonstrate that DeepProbLog supports 
\begin{enumerate*}[label=(\roman*)]
\item logical reasoning and deep learning;
\item program induction; and 
\item probabilistic inference and combined probabilistic and deep learning. 
\end{enumerate*}

We provide implementation details at the end of this section and list all programs in \ref{app:Programs}.
\subsection{Logical reasoning and deep learning}
To show that DeepProbLog supports both logical reasoning and deep learning, 
we extend the classic learning task on the MNIST dataset \citep{mnist98} to four more complex problems that require reasoning:
\begin{description}[leftmargin=0.75cm]

\item[T1:] $\mathtt{addition(\digit{mnist_3},\digit{mnist_5},8)}$\\
Instead of using labeled single digits, we  train on pairs of images, labeled with the sum of the individual labels. 
This is the same as Example~\ref{ex:addition}.
The DeepProbLog program consists of the clause 
\[\mathtt{addition(X,Y,Z) {:}{-} digit(X,X2), digit(Y,Y2), Z~is~X2+Y2}
\]and a neural AD for the \texttt{digit/2} predicate, which classifies an MNIST image.
We compare to a CNN baseline \footnote{We'd like to thank Paolo Frasconi for the interesting discussion and idea for a new baseline.} classifying the two images into the 19 possible sums.

\paragraph{Results} Figure~\ref{fig:mnist_single_addition} shows the learning curves for the baseline (orange) and DeepProbLog (blue) on the single-digit addition. We evaluated on 3 levels of data availability: 30 000 examples, 3 000 and 300 examples. As can be seen in the figures, DeepProbLog converges faster and achieves a higher accuracy than the baseline. In the case for N = 30 000 (Figure~\ref{fig:mnist1}), the difference between the baseline and DeepProbLog is significant, but not immense. However, for N = 3000 and especially N = 300, the difference becomes more apparent. \\
The reason behind this disparity is that the baseline needs to learn making a decision for the combined input digits (and there are a 100 different sums possible), whereas the DeepProbLog's neural predicate only needs to recognize individual digits (with only 10 possibilities). Table~\ref{tab:mnist_single_addition} shows the average accuracy on the test set for the different models for different training set sizes.

\begin{figure}[t]
    \centering
        \begin{subfigure}[b]{0.32\linewidth}
        \centering
        \includegraphics[width=\linewidth]{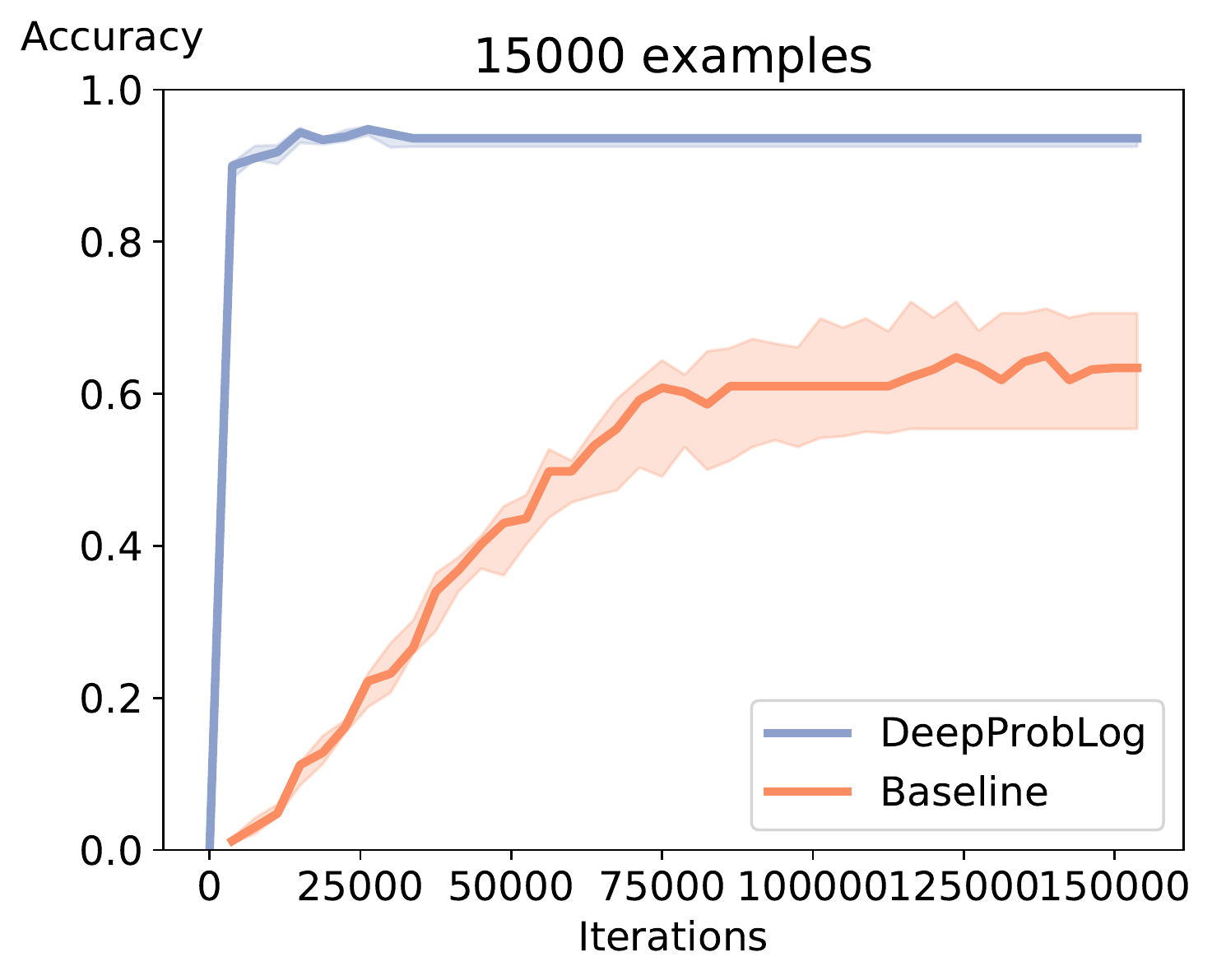}
        \caption{15 000 examples}
        \label{fig:multi_mnist1}
    \end{subfigure}
    \begin{subfigure}[b]{0.32\linewidth}
        \centering
        \includegraphics[width=\linewidth]{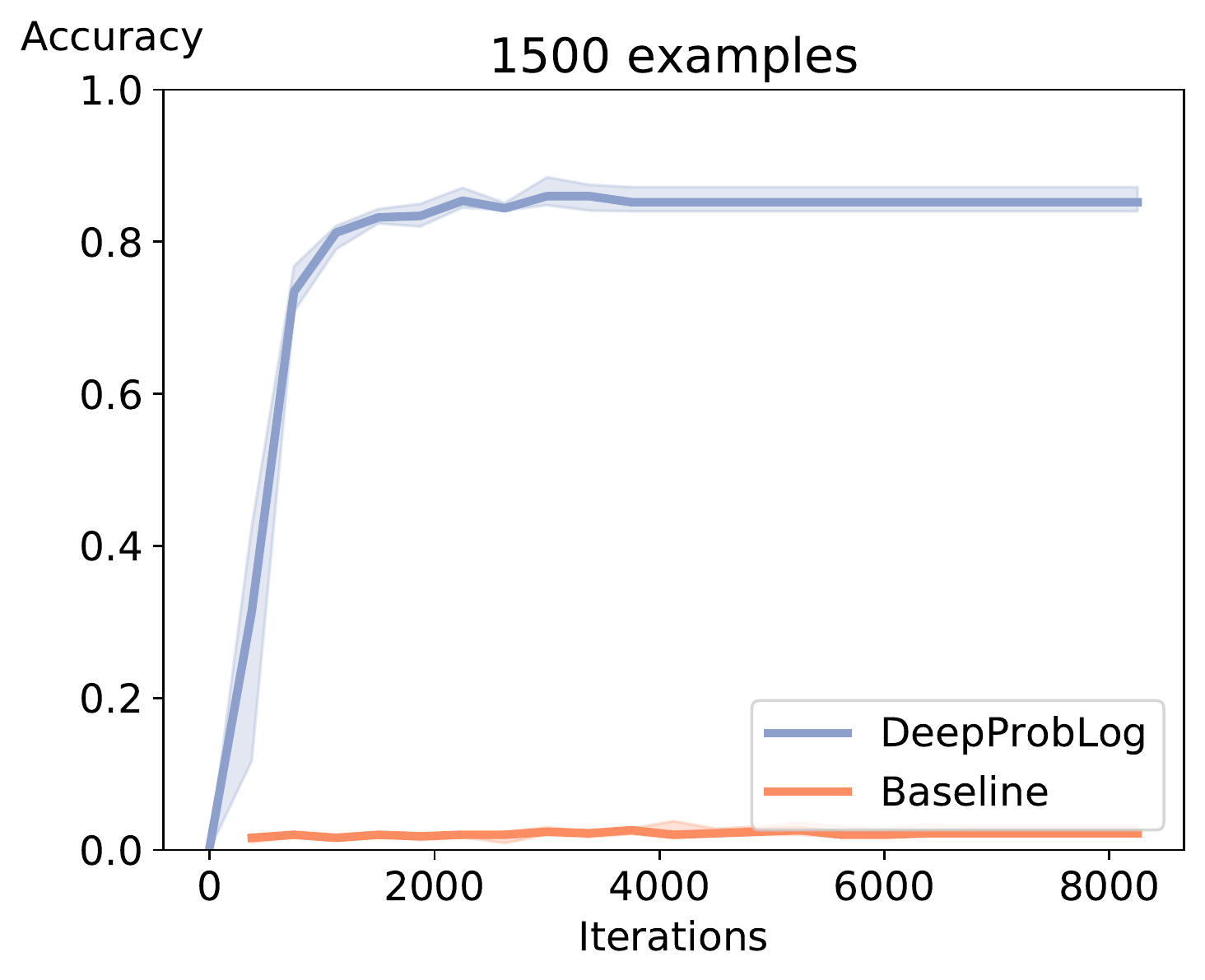}
        \caption{1 500 examples}
        \label{fig:multi_mnist2}
    \end{subfigure}
    \begin{subfigure}[b]{0.32\linewidth}
        \centering
        \includegraphics[width=\linewidth]{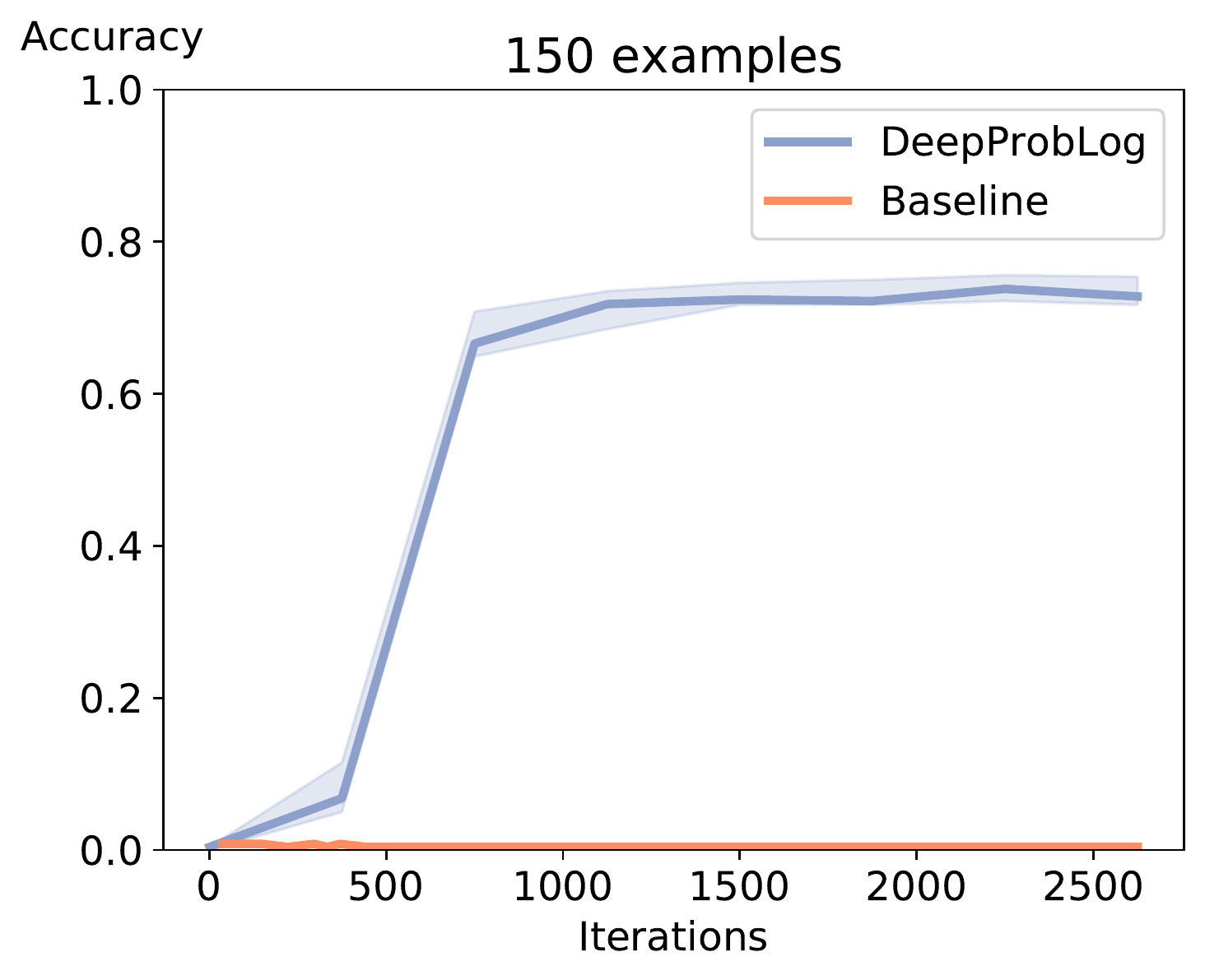}
        \caption{150 examples}
        \label{fig:multi_mnist3}
    \end{subfigure}
    \caption{MNIST Multi-Digit Addition (\textbf{T2}). The graphs show the accuracy on the validation set during training for different training set sizes.}
    \label{fig:mnist_multi_addition}
\end{figure}

\begin{table}[t]
    \centering
\begin{tabular}{@{}lrrrl@{}}
\toprule
            & \multicolumn{4}{c}{Number of training examples}                                \\ \cmidrule(l){2-5}
        Model       & 15 000          & 1 500          & 150    & \textbf{T1} (30 000)     \\ \midrule
        Baseline & $60.85 \pm 9.77$ & $1.34 \pm 0.53$ & $0.80 \pm 0.14$ & --\\
        DeepProbLog & $95.16 \pm 1.70$ & $87.21 \pm 1.92$ & $72.73 \pm 3.03$ & $93.36 \pm 1.18$ \\ \bottomrule
    \end{tabular}
    \caption{The accuracy on the test set for \textbf{T2}.}
    \label{tab:mnist_multi_addition}
\end{table}

\item[T2:] $\mathtt{addition([\digit{mnist_3},\digit{mnist_8}],[\digit{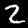},\digit{mnist_5}],63)}$\\
The  input consists of two lists of images, each element being a digit. Each list represents a multi-digit number. The label is the sum of the two numbers. The neural predicate remains the same.
Learning the new predicate requires only a small change in the logic program. Because the CNN baseline cannot handle numbers of varying size, we fixed the size of the input to two-digit numbers.
\paragraph{Results} First, we perform an experiment where we take the neural network trained in \textbf{T1} and use it in this model without any further training. Evaluating it on the same test set, we achieve an accuracy that is not significantly different from training on the full dataset of \textbf{T2}. This demonstrates that the approach used in DeepProbLog causes it to generalize well beyond training data.
Figure~\ref{fig:mnist_multi_addition} shows the learning curves for the baseline (orange) and DeepProbLog (blue) on the multi-digit addition. DeepProbLog achieves a somewhat lower accuracy compared to the single digit problem due to the compounding effect of the classification error on the individual digits, but the model generalizes well. The baseline fails to learn from few examples (150 and 1 500). It is able to learn with 15 000 examples, but converges very slowly. Table~\ref{tab:mnist_multi_addition} shows the average accuracy on the test set for the different models for different training set sizes.

\item[T3:] $\mathtt{addition(\digit{mnist_3},\digit{mnist_5},\digit{mnist_8})}$\\
The input consists of 3 MNIST images such that the last is the sum of the first two. This task demonstrates potential pitfalls of only providing supervision on the logic level. Namely, without any regularization, the neural network quickly learns to predict $0$ for all digits, i.e., the model collapses to always predicting $0+0=0$, as it is a valid logical solution. To avoid this, we add a regularisation term based on entropy maximization (Equation~\ref{eq:infoloss}, Section~\ref{sec:implementation}). The intuition behind this regularisation term is that it penalizes mode collapse by requiring the entropy of the average output distribution per batch to be high. As such, this term encourages exploration, but is only necessary to start the training of the neural networks. If they are sufficiently trained, this term can be dropped. We call this additional loss term \emph{infoloss}. This additional regularization loss is multiplied by a factor $\lambda$ and added to the cross-entropy loss. We run the experiment for different values of $\lambda$.

\paragraph{Results}Figure~\ref{fig:all_digit} shows the accuracy of the neural predicate on classifying single digits for different levels of the regularization parameter. As can be seen, for $\lambda = 2$, the neural predicate  converges on the trivial solution. For $\lambda = 4$, the neural predicate sometimes converges on the correct solution, but can also converge on the wrong solution. For $\lambda = 8$, the neural network consistently converges on the correct solution.

\item[T4:] $\mathtt{addition(\digit{mnist_3},\digit{mnist_5},\textcolor{red}{14})}$\\
This experiment is the example shown in Figure~\ref{fig:dpl_learning}. It's the same as \textbf{T1}, but with noise introduced in the labels. Namely, a fraction of the labels is replaced by uniformly selected number between $0$ and $18$. We compare three models: the CNN baseline from \textbf{T1}, the DeepProbLog model from \textbf{T1}, and a DeepProbLog model where the noise is explicitly modeled as in Figure~\ref{fig:dpl_learning}.
\paragraph{Results}Table~\ref{tab:noisy_addition} shows the accuracy on the test set which has no noise. The baseline is not tolerant to noisy labels, quickly dropping in accuracy as the fraction of noisy labels increases. The DeepProbLog model from \textbf{T1} is more tolerant, but also drops noticeably in accuracy as the fraction of noise goes over $0.5$. Explicitly modeling the noise makes the model very noise tolerant, even retaining an accuracy of $73.2\%$ with $80\%$ noisy labels. As shown in the last row, it is also able to learn the fraction of noisy labels in the data. This shows that the model is able to recognize which examples have noisy labels.
\end{description}

\begin{figure}[t]
    \centering
    \includegraphics[width=0.45\linewidth]{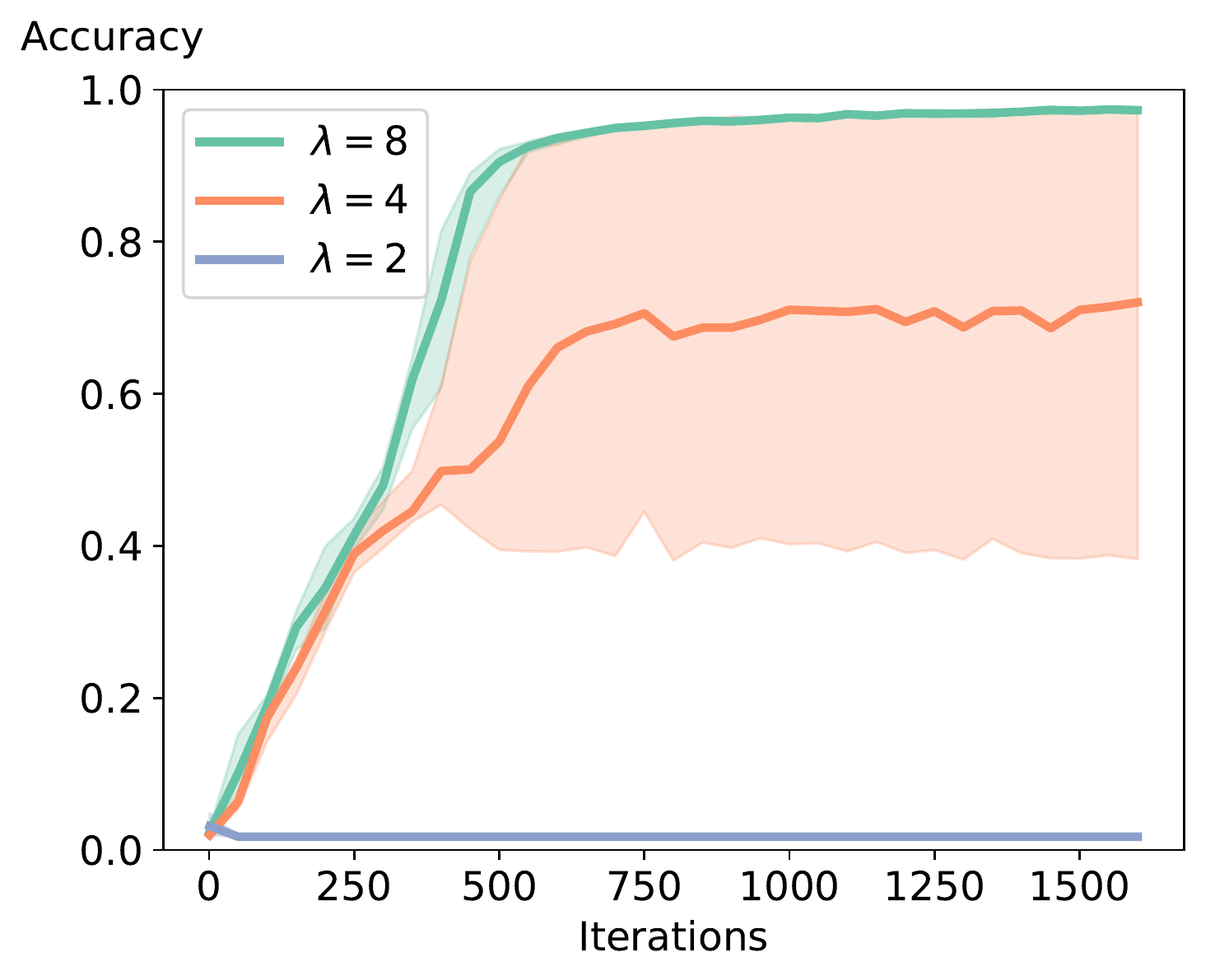}
    \caption{The accuracy on the MNIST test set for individual digits while training on (\textbf{T3}).}
    \label{fig:all_digit}
\end{figure}

\begin{table}[t!]
\centering
\begin{tabular}{@{}lrrrrrr@{}}
\toprule
                              & \multicolumn{6}{c}{Fraction of noise}          \\ 
                              & 0.0   & 0.2    & 0.4   & 0.6   & 0.8   & 1.0   \\ \midrule
Baseline                      & $93.46$ & $87.85$ & $82.49$ & $52.67$ & $8.79$ & $5.87$     \\
DeepProbLog                   & $97.20$ & $95.78$ & $94.50$ & $92.90$ & $46.42$ & $0.88$  \\ \midrule
DeepProbLog w/ explicit noise & $96.64$ & $95.96$ & $95.58$ & $94.12$ & $73.22$ & $2.92$ \\ 
Learned fraction of noise     & $0.000$ & $0.212$ & $0.415$ & $0.618$ & $0.803$ & $0.985$ \\
\bottomrule
\end{tabular}
\caption{The accuracy on the test set for \textbf{T4}.}
\label{tab:noisy_addition}.
\end{table}

\subsection{Program Induction}
The second set of problems demonstrates that DeepProbLog can perform program induction. 
We follow the program sketching \citep{sketching} setting of differentiable Forth ($\partial 4$) \citep{riedel2017programming}, where holes in given programs need to be filled by neural networks trained on input-output examples for the entire program. As in their work, we consider three tasks: addition, sorting \citep{reed2015neural} and word algebra problems (WAPs) \citep{roy2016solving}. 

\begin{description}[leftmargin=0.75cm]
\item[T5:] $\mathtt{forth\_addition([4],[8],1,[1,3])}$\\
The input consists of two numbers, represented as lists of digits, and a carry. The output is the sum of the numbers and the  carry.
The program specifies the basic addition algorithm in which we go from right to left over all digits, calculating the sum of two digits and taking the carry over to the next pair. The hole in this program corresponds to calculating the resulting digit (\texttt{result/4}) and carry (\texttt{carry/4}), given two digits and the previous carry. 
\paragraph{Results} The results are shown in Table~\ref{tab:forth_addition}. Similarly to $\partial 4$ , DeepProbLog achieves 100\% on all training sizes.

\item[T6:] $\mathtt{forth\_sort([8, 2, 4],[2, 4, 8])} $\\
The input consists of a list of numbers, and the output is the sorted list.  The program implements bubble sort, but leaves open what to do on each step in a bubble (i.e. whether to swap or not: \texttt{swap/2}).
\paragraph{Results}The results are shown in Table~\ref{tab:forth_sort}. Similarly to $\partial 4$ , DeepProbLog achieves 100\% on training sizes 2 and 3. However, whereas $\partial 4$ fails to converge on training sizes larger than 3, DeepProbLog stills achieves 100\% accuracy. As \citet{riedel2017programming} mention, the failure of $\partial 4$ is due to computational issues arising from the long program trace resulting from sorting long lists. DeepProbLog does not suffer from these issues.
As shown in Table~\ref{tab:timing}, DeepProbLog runs faster and scales better with increasing training length.

\item[T7:]\texttt{wap(}\emph{`Robert has 12 books . ... How many does he have now ?'}\texttt{,12,3,1,10)}\\
The input to the WAPs consists of a natural language sentence describing a simple mathematical problem. These WAPs always contain three numbers, which are extracted from the string and are given as part of the input. The output is the solution to the question.  Every WAP can be solved by chaining the following  4 steps: permuting the three numbers (\texttt{permute/2}), applying an operation on the first two numbers (addition, subtraction or product \texttt{operation\_1/2}), potentially swapping the intermediate result and the last digit (\texttt{swap/2}), and performing a last operation (\texttt{operation\_2/2}). The hole in the program is in deciding which of the alternatives should happen on each step.
\paragraph{Results}DeepProbLog reaches an accuracy of up to 96.5\%, similar to  the results for $\partial 4$ reported by \citet{riedel2017programming} (96\%).
\end{description}
\begin{table}[t]
\centering
\begin{tabular}{@{}lrrrr@{}}
\toprule
                                                            &             & \multicolumn{3}{c}{Training length} \\ \cmidrule(l){3-5} 
                                                            & Test length & 2          & 4          & 8         \\ \midrule
\multirow{2}{*}{$\partial 4$ \citep{riedel2017programming}} & 8           & 100.0      & 100.0      & 100.0     \\
                                                            & 64          & 100.0      & 100.0      & 100.0     \\ \midrule
\multirow{2}{*}{DeepProbLog}                                & 8           & 100.0      & 100.0      & 100.0     \\
                                                            & 64          & 100.0      & 100.0      & 100.0     \\ \bottomrule
\end{tabular}
\caption{Accuracy on the addition (\textbf{T5}) problem (results for $\partial 4$ reported by \citet{riedel2017programming}).}
\label{tab:forth_addition}
\end{table}

\begin{table}[t]
\centering
\begin{tabular}{@{}lrrrrrr@{}}
\toprule
                                                                             &             & \multicolumn{5}{c}{Training length}   \\ \cmidrule(l){3-7} 
                                                                             & Test length & 2     & 3     & 4     & 5     & 6     \\ \midrule
\multirow{2}{*}{$\partial 4$ \citep{riedel2017programming}} & 8           & 100.0 & 100.0 & 49.22 & --    & --    \\
                                                                             & 64          & 100.0 & 100.0 & 20.65 & --    & --    \\ \midrule
\multirow{2}{*}{DeepProbLog}                                                 & 8           & 100.0 & 100.0 & 100.0 & 100.0 & 100.0 \\
                                                                             & 64          & 100.0 & 100.0 & 100.0 & 100.0 & 100.0 \\ \bottomrule
\end{tabular}
\caption{Accuracy on the sorting (\textbf{T6}) problem (results for $\partial 4$ reported by \citet{riedel2017programming}).}
\label{tab:forth_sort}
\end{table}

\begin{table}[t]
\centering
\begin{tabular}{@{}lrrrrr@{}}
\toprule
                    & \multicolumn{5}{c}{Training length} \\ \cmidrule(l){2-6} 
                    & 2    & 3     & 4    & 5     & 6     \\ \midrule
$\partial 4$ on GPU & 42 s & 160 s & --   & --    & --    \\
$\partial 4$ on CPU & 61 s & 390 s & --   & --    & --    \\
DeepProbLog         & 11 s & 14 s  & 32 s & 114 s & 245 s \\ \bottomrule
\end{tabular}
\caption{Time until 100\% accurate on test length 8 for the sorting (\textbf{T6}) problem.}
\label{tab:timing}
\end{table}

\newpage
\subsection{Probabilistic programming and deep learning}\label{subsec:probprogr}

In this section we introduce two final experiments that show the intricacies involved in combining probabilistic logic programming and deep learning.

\begin{description}[leftmargin=0.75cm]
\item[T8:] \textit{Coin classification and comparison}\\
In this experiment we train two neural networks using distant supervision. The input consists of a synthetic image containing two coins (an example is shown in Figure~\ref{fig:t8}.
They are either heads or tails. The image is labeled either with \emph{same} or \emph{different}. We train a neural network for each coin to predict either \emph{heads} or \emph{tails}. 
Solving this task requires solving two problems. On the one hand, the neural networks have to learn to recognize and separate the two different coins; on the other hand, they also have to each classify a different coin as heads or tails.
The first question we ask is whether the neural networks can recover the latent structure imposed by the logic program. We expect the two neural networks to agree on which side of the coin is heads and which is tails, however, this might be the inverse of what is generally considered heads and tails. Furthermore, we expect the two neural networks to each pick one coin to label, but which network classifies which coin will vary between runs. As such, there are four possible solutions that the neural networks can converge on. The second question we ask is how many additionally labeled examples (with both the label for same/different and heads/tails of one of the coins given) we need for the neural network to recover the desired latent  representation.
\paragraph{Results}
We ran each experiment 100 times. The fraction of runs that converged on either no solution, the expected solution or a logically equivalent solution is shown in Table~\ref{tab:t8}. We see that with no additionally labeled examples, DeepProbLog doesn't converge on a satisfactory solution in about half of all runs. When it does converge on a solution, it converges on the \emph{expected} solution 25\% of the time, and on different solutions 75\% of the time, which is conform with our expectations.
We can also see that as the number of additionally labeled examples increases, DeepProbLog converges more reliably, and more on the expected solution. Starting with 10 additionally labeled examples, DeepProbLog reliably converges on the desired solution. Beyond 20 additionally labeled examples, we don't see any further improvements.
\begin{table}[t]
    \centering
\begin{tabular}{@{}rrrr@{}}
\toprule
Labeled examples & Not solved & Expected solution & Other solution \\ \midrule
0                & 56\%      & 11\%             & 33\%             \\
5                & 39\%          & 40\%             & 21\%              \\
10               & 7\%          & 92\%                & 1\%            \\
20               & 4\%         & 96\%                & 0\%             \\ 
50               & 3\%          & 97\%              &0\%\\
100               & 4\%         & 96\%                & 0\%             \\ \bottomrule
\end{tabular}
    \caption{The fraction of runs that converged to different outcomes for the Coins experiment (\textbf{T8}).}
    \label{tab:t8}
\end{table}
\begin{figure}[t]
    \centering
    \includegraphics[width=0.5\linewidth]{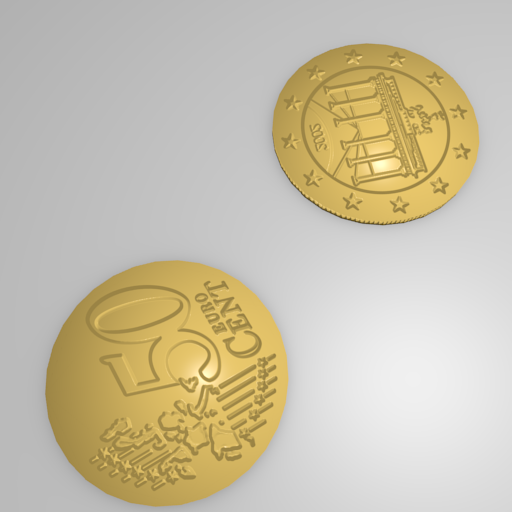}
    \caption{An example input image for the Coins (\textbf{T8}) experiment.}
    \label{fig:t8}
\end{figure}
\item[T9:] \texttt{0.8::poker([\textcolor{red}{Q$\heartsuit$}, \textcolor{red}{Q$\diamondsuit$}, \textcolor{red}{A$\diamondsuit$}, K$\clubsuit$],loss).}\\
In this last experiment we demonstrate that DeepProbLog can perform combined probabilistic reasoning, probabilistic learning and deep learning. We do this by playing a simplified poker game: there are two players that are dealt two cards from several decks. There is also one community card. Each player then makes a poker hand (e.g. pair, straight, ...) with their two cards and the community card. 

For simplicity, we only use the jack, queen, king and ace. We also do not consider the suits of the cards.

The input consists of 4 images that show the cards dealt to the two players.  Additionally, every example is labeled with the chance that the game is won, lost or ended in a draw, e.g.:
\[
\mathtt{0.8::poker([\textcolor{red}{Q\heartsuit}, \textcolor{red}{Q\diamondsuit}, \textcolor{red}{A\diamondsuit}, K\clubsuit],loss)}
\]
We expect DeepProbLog to:
\begin{itemize}
    \item train the neural network to recognize the four cards
    \item reason probabilistically about the non-observed card
    \item learn the distribution of the unlabeled community card
\end{itemize}
To make DeepProbLog converge more reliably, we add some examples with additional supervision. Namely, in 10\% of the examples we additionally specify the community card, i.e.
\[
\mathtt{poker([\textcolor{red}{Q\heartsuit}, \textcolor{red}{Q\diamondsuit}, \textcolor{red}{A\diamondsuit}, K\clubsuit],\textcolor{red}{A\diamondsuit},loss)}.
\]
This also showcases one of the strengths of DeepProbLog, namely, it can make use of examples that have different levels of observability. 
The loss function used in this experiment is the MSE between the predicted and target probabilities.

\paragraph{Results}
We ran the experiment 10 times. Out of these 10 runs, 4  didn't converge on the correct solution. The average values of the learned parameters for the remaining 6 runs are shown Table~\ref{tab:t9}. As can be seen, DeepProbLog is able to correctly learn the probabilistic parameters. In these 6 runs, the neural network also correctly learned to classify all card types, achieving a 100\% accuracy. The other runs did not converge because some of the classes were permuted (i.e., queens predicted as aces and vice versa) or multiple classes mapped onto the same one (queens and kings were both predicted as kings).
\begin{table}[t]
    \centering
    \begin{tabular}{@{}lrrrr@{}}
    \toprule
     Distribution  & Jack & Queen & King & Ace  \\ \midrule
    Actual  & 0.2  & 0.4   & 0.15 & 0.25 \\
    Learned &  $0.203\pm 0.002$    & $0.396\pm 0.002$       &$0.155 \pm 0.003$& $0.246 \pm 0.002$ \\ \bottomrule
    \end{tabular}
    \caption{The results for the Poker experiment (\textbf{T9}).}
    \label{tab:t9}
\end{table}
\end{description}

\subsection{Implementation details}
\label{sec:implementation}
For the implementation we integrate ProbLog2 \citep{dries2015problog2} with PyTorch \citep{paszke2017automatic}. All programs are listed in \ref{app:Programs}. In experiments \textbf{T1-T8} we optimize the cross-entropy loss between the predicted and target query probabilities, as we found that this works better to learn the probabilistic parameters. In experiment \textbf{T9} we optimize the MSE between the predicted and target query probabilities. We use Adam \citep{kingma2014adam} optimization for the neural networks, and SGD for the logic parameters. For \textbf{T9}, we add random rotations (max 10 degrees) and shift the colours in the HSV by up to 5\% to the images of the cards. \\
The neural network architectures are summarized in Table~\ref{tab:architectures}. \textit{Conv(o,k)} denotes a convolutional layer with o output channels and kernel size k. \textit{Lin(n)} denotes a fully connected layer of size n. \textit{BiGRU(h)} denotes a single-layer bi-directional GRU with a hidden size h. \textit{(layer)$\times 2$} means that there are two identical layers in parallel that are concatenated. A layer in bold means it is followed by a ReLU activation function. All neural networks end with a Softmax layer, unless otherwise specified. The hyperparameters used in the experiments are shown in Table~\ref{tab:parameters}. The sizes of the datasets used are specified in Table~\ref{tab:dataset_sizes}.\\

The regularisation term used in \textbf{T3} is calculated per network and per batch on the average of the neural network output. It is calculated as
\begin{equation}\label{eq:infoloss}
1.0 - H_n\left(\frac{1}{N}\sum_{i=1}^n P_i\right)
\end{equation}
where $P_i$ is the i-th output of the neural network and $H_n$ is the n-ary entropy (i.e. entropy using the base-n logarithm).

\subsection{Computation time}
Due to the nature of the exact inference used in DeepProbLog, which it inherits from aProbLog \cite{kimmig2011algebraic}, the grounding and compilation steps can become expensive as the problem size grows. For example, in \textbf{T1}, which is the smallest problem we consider, grounding and compilation takes on average 0.01 seconds, while evaluating the NN and AC takes on average 0.002 seconds per example. In \textbf{T9}, which has the largest logic program out of all experiments, grounding and compilation takes on average 1.3 seconds,  while NN and AC evaluation takes on average 0.05 seconds per example.

It is important to note that when we evaluate an example a second time, the structure of the AC, which is determined by the grounding and compilation, remains the same. Only the learned probabilities  in the nAD change. We make use of this to improve the performance by caching the arithmetic circuits so that we only have to perform the (potentially expensive) grounding and compilation steps once. During evaluation, we only re-evaluate the neural networks and evaluate the AC with the updated probabilities.

Note that this optimization can also be applied to, for example, the queries $\mathtt{addition(\digit{mnist_3},\digit{mnist_5},8)}$ and $\mathtt{addition(\digit{mnist_8},\digit{0},8)}$, as both have the same structure. To do this, we introduce placeholder constants and change both queries to the single query $\mathtt{addition(a,b,8)}$, which reduces all queries in \textbf{T1} to 19 unique queries, one for each different label. During the evaluation of the neural networks, we replace the constants with the correct input and use the resulting probabilities in the cached ACs. We apply these optimizations to all experiments.

\begin{table}[t]
\centering
\begin{tabular}{@{}lllll@{}}
\toprule
Task           & Batch size & Learning rate & Parameter learning rate & Infoloss \\ \midrule
\textbf{T1-T3} & 2         & 1e-3          &                         &          \\
\textbf{T4}    & 2         & 1e-3          & 1e-3                    & 2, 4, 8  \\ \midrule
\textbf{T5}    & 50          & 0.02          &                         &          \\
\textbf{T6}    & 16         & 1.0           &                         &          \\
\textbf{T7}    & 100        & 0.005         &                         &          \\ \midrule
\textbf{T8}    & 5          & 1e-4          &                         & 0.25     \\
\textbf{T9}    & 50         & 1e-4          & 1e-3                    & 0.5      \\ \bottomrule
\end{tabular}
\caption{Overview of the hyperparameters used in the experiments.}
\label{tab:parameters}
\end{table}

\begin{table}[t]
\centering
\begin{tabular}{@{}lrrr@{}}
\toprule
Task        & Training set        & Validation Set & Test set \\ \midrule
\textbf{T1} & 29 500,  3 000, 300 & 500            & 5 000    \\
\textbf{T2} & 14750, 1 500, 150   & 250            & 2 500    \\
\textbf{T3} & 16 000              & 2 000          & 3 000    \\
\textbf{T4} & 29 500              & 500            & 5 000    \\ \midrule
\textbf{T5} & 512                 & 256            & 1 024    \\
\textbf{T6} & 256                 & 32             & 32       \\
\textbf{T7} & 300                 & 100            & 200      \\ \midrule
\textbf{T8} & 100                 & --             & 20       \\
\textbf{T9} & 500                 & --             & 25       \\ \bottomrule
\end{tabular}
\caption{Overview of the sizes of the datasets used in the experiments.}
\label{tab:dataset_sizes}
\end{table}

\begin{table}[t]
\centering

\begin{tabular}{@{}lll@{}}
\toprule
Task           & Network  & Architecture                                                                                                                                 \\ \midrule
\textbf{T1-T4} & digit/2  &  MNISTConv,  \textbf{Lin(120)}, \textbf{Lin(84)}, Lin(10)                       \\
\textbf{T1,T3} & baseline & MNISTConv$\times 2$, \textbf{Lin(120)}, \textbf{Lin(84}, Lin(19)  \\
\textbf{T2}    & baseline & (MNISTConv$\times 2$, \textbf{Lin(100)})$\times 2$,\textbf{Lin(128}, Lin(199) \\ \midrule
\textbf{T5}    & result/4 & Lin(50), TanH, Lin(10)                                                                                                                       \\
               & carry/4  & Lin(10), TanH, Lin(2)                                                                                                                        \\
\textbf{T6}    & swap/3   & \textbf{Lin(20)}, Lin(10)                                                                                                                    \\
\textbf{T7}    & RNN      & Embedding(256), BiGRU(512), Dropout(0.5)*                                                                                   \\
               & perm/2   & Lin(6)                                                                                                                                       \\
               & op1/2    & Lin(4)                                                                                                                                       \\
               & swap/2   & Lin(2)                                                                                                                                       \\
               & op2/2    & Lin(4)                                                                                                                                       \\ \midrule
\textbf{T8}    & coin1/2  & AlexNetConv, \textbf{Lin(100)}, Lin(2)                                                                                                       \\
               & coin2/2  & AlexNetConv, \textbf{Lin(100)}, Lin(2)                                                                                                       \\
\textbf{T9}    & rank/2   & AlexNetConv, \textbf{Lin(100)}, Lin(4)                                                                                                       \\ \bottomrule
\end{tabular}\\
{\small 
MNISTConv:  Conv(6,5), \textbf{MP(2,2)}, Conv(16,5), \textbf{MP(2,2)}*\\
AlexNetConv: \textbf{Conv(64, 11, 2,2)}, MP(3,2), \textbf{Conv(192, 5, 2)}, MP(3,2), \\\textbf{Conv(384, 3, 1)}, \textbf{Conv(256, 3, 1)}, \textbf{Conv(256, 3, 1)}, MP(3,2)*}\\
* Does not end with a Softmax layer.
\caption{Overview of the neural network architectures used in the experiments.}
\label{tab:architectures}
\end{table}

\section{Related Work}

Most of the work on combining neural networks and logical reasoning comes from the \textit{neuro-symbolic reasoning} literature \citep{garcez2012neural,hammer2007perspectives}. 
These approaches typically focus on approximating logical reasoning with neural networks by encoding logical terms in Euclidean space.
However, they neither support probabilistic reasoning nor perception, and are often limited to 
non-recursive and acyclic logic programs \citep{Holldobler1999}. 
DeepProbLog takes a different approach and integrates neural networks into a probabilistic logic framework, retaining the full power of both logical and probabilistic reasoning and deep learning.

At the same time, DeepProblog also 
integrates probability  in neuro-symbolic computation. Although this may appear as a complication, our work actually shows that it  can greatly simplify the integration of neural networks with logic. 
The reason for this is that the probabilistic framework provides a clear optimisation criterion, namely the probability of the training examples. Real-valued probabilistic quantities are also well-suited for gradient-based training procedures, as opposed to discrete logic quantities.

The prominent recent lines of related work focus on three main branches: pushing the logic as regularisation, templating neural networks, and neural program induction.

\subsection{Logic as regularisation} 
The main idea behind this line of research is that logic is included as a regularizer during the optimization of the neural network, or the learning of the embeddings. The goal is to encode the logic into the weights so that after training, when the logic is no longer explicitly present, the evaluation still shows the characteristics of the logic. 
\cite{rocktaschel2015injecting, Demeester2016, minervini2017adversarial,  diligenti2017semantic,donadello2017logic,xu2018semantic}
all center around including logical background knowledge as a regularizer during training. 
\citet{rocktaschel2015injecting} inject background knowledge into a matrix factorization model for relation extraction, by adding differentiable loss terms for propositionalized first-order rules. \citet{Demeester2016} propose a more efficient alternative by inducing order relations in embedding space, effectively leading to a lifted application of the rules. This is further generalized by \citet{minervini2017adversarial}, who investigate injecting rules by minimizing an inconsistency loss on adversarially-generated examples.
\citet{diligenti2017semantic} use FOL to specify constraints on the output of the neural network. They use fuzzy logic to create a differentiable way of measuring how much the output of the neural networks violates these constraints. This is then added as an additional loss term that acts as a regularizer. More recent work by \citet{xu2018semantic} introduces a similar method that uses probabilistic logic instead of fuzzy logic, and is thus more similar to DeepProbLog. They also compile the formulas to an SDD for efficiency.

However, whereas DeepProbLog is based on (probabilistic) logic programming, these methods use first order logic instead. This is reminiscent to the difference between ProbLog and Markov Logic \cite{richardson2006markov} or PSL \cite{bach2015hinge}.
\citet{donadello2017logic}, though at first sight related to \citet{diligenti2017semantic}, work slightly differently. They learn functions that map numerical properties of logical constants onto truth values, which are then combined using fuzzy logic.

\subsection{Templating neural networks} \label{subsec:templating}

This line of work uses the logic as a template for constructing the architecture of neural networks.  This is reminiscent of the knoweldge base construction approach of statistical relational artificial intelligence \cite{DeRaedt16}.

\citet{rocktaschel2017end}  introduce a differentiable framework for theorem proving. 
They re-implemented Prolog's theorem proving procedure in a differentiable manner and enhanced it with learning subsymbolic representation of the existing symbols, which are used to handle  noise in data. 
Whereas \citeauthor{rocktaschel2017end} use logic only to construct a neural network and focus on learning subsymbolic representations, DeepProblog focuses on tight interactions between the two and parameter learning for both the neural  and  the logic components. In this way, DeepProbLog retains the best abilities of both worlds. Recently, \citet{weber2019nlprolog} extend the notion of soft 
unification towards structured textual knowledge, i.e., unification can 
be performed between sentences, not only symbols. In contrast to 
\citet{rocktaschel2017end}, \citet{weber2019nlprolog} retain the full ability of
logical reasoning, and as such is closer to DeepProblog, but it is
specialised for  NLP tasks.

\citet{cohen2018tensorlog} introduce a framework to compile a tractable subset of logic programs into differentiable functions and to execute it with neural networks.  
It provides an alternative probabilistic logic but it has a different and less developed semantics. Furthermore, to the best of our knowledge it has not been applied to the kind of tasks tackled in the present paper.
An idea similar in spirit to ours is that of \citet{andreas2016neural}, who introduce a neural network for visual question answering composed out of smaller modules responsible for individual tasks, such as object detection. Whereas the composition of modules is determined by the linguistic structure of the questions, DeepProbLog uses probabilistic logic programs to connect the neural networks. 

\subsection{Neural program induction.}

The third line of work has focused on learning programs from data by combining neural and symbolic approaches.

\paragraph{Neural execution}
The first category captures a program behaviour with neural networks and therefore focuses on program execution.
The approach most similar to ours is that of \citet{riedel2017programming}, where neural networks are used to fill in \textit{holes} in a partially defined Forth program. 
DeepProblog differs in that it uses ProbLog as the host language which results in native support for both logical and probabilistic reasoning, while differentiable Forth uses a procedural language.
Differentiable Forth has been applied to tasks T5-7, but it is unclear whether it could be applied to the remaining ones.
Finally, \citet{evans2018learning} introduce a differentiable framework for rule induction, that does not focus on the integration of the two approaches like DeepProblog.

\paragraph{Neurally guided search}
The second line of research enhances the search procedures of the symbolic program induction techniques by incorporating neural components in the search itself.
The key principle these techniques employ is to perform the search over programs in a systematic symbolic way, but guide the search with a heuristic learned by a deep neural network.
For instance, \citet{kalyan2018} train a neural network to predict the scores of branches during the branch-and-bound search procedure, \citet{zhang2018neural} train a neural network to choose which candidate program to expand next while exploiting the constraints on the input-output examples, while \citet{Ellis2018} use a neural network to efficiently search over a well-designed DSL.

\paragraph{Neural program construction}
The final category involves techniques that decompose a problem into independent parts that can be individually solved by either neural or symbolic components and synchronize the individual components to solve the main problem.
For instance, \citet{Yi2018,mao2018} develop a neuro-symbolic approach towards visual question answering by using a neural network to generate a program computing the answer to the question and executing the program symbolically.
\citet{Ellis2018a} generate a \LaTeX{} code from a hand-drawn sketch by using a neural network to recognise basic shapes within a sketch and symbolically inducing the program describing the sketch.
In contrast, \citet{dong2018neural} induce programs in a purely neural way and demonstrate favourable generalization; however, they do not induce symbolic programs, but rather express a program through a neural network.

\subsection{Symbolic deep learning}
The success of neural deep learning has inspired has inspired several works introducing symbolic deep learning methods which, instead of representing the logical aspects in a vector space, retain the logical data representation in the latent representation.
These include the symbolic versions of deep neural networks: \citet{LRNN} treat symbolic rules expressed in first-order logic as a template for constructing a neural network, while \citet{KazemiAAAI18} compose a relational neural network by adding hidden layers to relational logistic regression \cite{Kazemi:2014:RLR:2908339.2908347}.
Another research direction focuses on task-agnostic discovery of relational (symbolic) latent representations by exploiting approximate symmetries \cite{DumancicIjcai2017}, a symbolic extension of the auto-encoding principle \cite{DumancicIjcai2019}, or self-play \cite{CropperIjcai2019}.

\section{Conclusion}
We introduced DeepProbLog, a framework where neural networks and probabilistic logic programming are integrated in a way that exploits the full expressiveness and strengths of both worlds  and can be trained end-to-end based on examples. This was accomplished by extending an existing probabilistic logic programming language, ProbLog, with neural predicates. Learning is performed by using aProbLog to calculate the gradient of the loss which is then used in standard gradient-descent based methods to optimize parameters in both the probabilistic logic program and the neural networks.
We evaluated our framework on experiments that demonstrate its capabilities in combined symbolic and subsymbolic  reasoning, program induction, and probabilistic logic programming.

Although we have shown promising results, DeepProbLog is currently using only exact inference. As exact inference does not always scale well and can be prohibitively expensive for large problems, the DeepProbLog implementation cannot yet be applied to problems such as KB completion. Future work will be concerned with incorporating   approximate inference algorithms to speed-up the grounding and compilation process.
\section*{Acknowledgements}
RM is a SB PhD fellow at FWO (1S61718N).
SD is supported by the Research Fund KU Leuven (GOA/13/010) and Research Foundation - Flanders (G079416N).
LDR is partially supported by the European Research Council Advanced Grant project SYNTH (ERCAdG-694980). 

\bibliography{bibliography.bib}
\newpage
\appendix
\section{DeepProbLog Programs}
\label{app:Programs}
\begin{lstfloat}[ht]
\footnotesize
\begin{minted}[frame=lines]{Prolog}
nn(m_digit,[X],Y,[0,...,9]) :: digit(X,Y).

addition(X,Y,Z) :- digit(X,X2), digit(Y,Y2), Z is X2+Y2.
\end{minted}
\caption{Single-digit MNIST addition (\textbf{T1})}
\label{fig:mnist-addition-program}
\end{lstfloat}

In Listing~\ref{fig:mnist-addition-program}, \texttt{digit/2} is the neural predicate that classifies an MNIST image into the integers 0 to 9. The \texttt{addition/3} predicate's first two arguments are MNIST digits, and the last is the sum. It classifies both images using \texttt{digit/2}  and calculates the sum of the two results.

\begin{lstfloat}[ht]
\footnotesize
\begin{minted}[frame=lines]{Prolog}
nn(m_digit,[X],Y,[0,...,9]) :: digit(X,Y).

number([],Result,Result).
number([H|T],Acc,Result) :-
    digit(H,Nr),
    Acc2 is Nr+10*Acc,
    number(T,Acc2,Result). 
number(X,Y) :- number(X,0,Y).

multi_addition(X,Y,Z) :- number(X,X2), number(Y,Y2), Z is X2+Y2.
\end{minted}
\caption{Multi-digit MNIST addition (\textbf{T2})}
\label{fig:multi-digit}
\end{lstfloat}

In Listing~\ref{fig:multi-digit}, the only difference with Listing~\ref{fig:mnist-addition-program} is that the \texttt{multi\_addition/3} predicate now uses the \texttt{number/2} predicate instead of the \texttt{digit/2} predicate. The \texttt{number/3} predicate's first argument is a list of MNIST images. It uses the \texttt{digit/2} neural predicate on each image in the list, summing and multiplying by ten to calculate the number represented by the list of images (e.g. \texttt{number([\includegraphics[height=7pt]{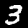},\includegraphics[height=7pt]{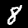}],38)}).

\begin{lstfloat}[ht]
\footnotesize
\begin{minted}[frame=lines]{Prolog}
nn(m_digit,[X],Y,[0,...,9]) :: digit(X,Y).

addition(X,Y,Z) :- digit(X,X2), digit(Y,Y2), digit(Z,Z2), Z2 is X2+Y2.
\end{minted}
\caption{All-digit MNIST addition (\textbf{T3})}
\label{lst:t3}
\end{lstfloat}

In Listing~\ref{lst:t3}, the only difference with Listing~\ref{fig:mnist-addition-program} is that  all 3 inputs \texttt{X,Y,Z} are images. As such, the \texttt{digit/2} predicate is also used on the third input. The sum is also redefined as \texttt{Z2 is X2+Y2}.

\begin{lstfloat}[ht]
\footnotesize
\begin{minted}[frame=lines]{Prolog}
nn(classifier, [X], Y, [0 .. 9]) :: digit(X,Y).
t(0.2) :: noisy.

1/19 :: uniform(X,Y,0) ; ... ; 1/19 :: uniform(X,Y,18).

addition(X,Y,Z) :- noisy, uniform(X,Y,Z).
addition(X,Y,Z) :- \+noisy, digit(X,N1), digit(Y,N2), Z is N1+N2.
\end{minted}
\caption{Noisy MNIST addition (\textbf{T4})}
\label{lst:t4}
\end{lstfloat}

In Listing~\ref{lst:t4}, an additional probabilistic fact (\texttt{noisy/2}) is added that encodes the chance of an example being noisy. 
The \texttt{addition/3} predicate is split into two cases: when the noisy is true or when noisy is false. The latter is the same as in Listing~\ref{fig:mnist-addition-program}. If \texttt{noisy} is true, Z is considered to be drawn from the uniform distribution (\texttt{uniform/3}).

\begin{lstfloat}[ht]
\footnotesize
\begin{minted}[frame=lines]{Prolog}
nn(m_result,[D1,D2,Carry],Y,[0,...,9])::result(D1,D2,Carry,Y).
                                
nn(m_carry,[D1,D2,Carry],Y,[0,1])::carry(D1,D2,Carry,Y).

slot(I1,I2,Carry,NewCarry,Result) :-
    result(I1,I2,Carry,Result),
    carry(I1,I2,Carry,NewCarry).
    
add([],[],[C],C,[]).

add([H1|T1],[H2|T2],C,Carry,[Digit|Res]) :-
    add(T1,T2,C,NewCarry,Res),
    slot(H1,H2,NewCarry,Carry,Digit).

forth_addition(L1,L2,C,[Carry|Res]) :- add(L1,L2,C,Carry,Res).
\end{minted}
\caption{Forth addition sketch (\textbf{T5})}
\label{lst:forth-addition}
\end{lstfloat}

In Listing~\ref{lst:forth-addition}, there are two neural predicates: \texttt{result/4} and \texttt{carry/4}. These are used in the \texttt{slot/4} predicate that corresponds to the slot in the Forth program. The first three arguments are the two digits and the previous carry to be summed. The next two arguments are the new carry and the new resulting digit. The \texttt{add/5} predicate's  arguments are: the two list of input digits, the input carry, the resulting carry and the resulting sum. It recursively calls itself to loop over both lists, calling the \texttt{slot/5} predicate on each position, using the carry from the previous step.

\begin{lstfloat}[ht]
\footnotesize
\begin{minted}[frame=lines]{Prolog}
nn(m_swap, [X]) :: swap(X,Y,).

hole(X,Y,X,Y):-\+swap(X,Y).

hole(X,Y,Y,X):-swap(X,Y).

bubble([X],[],X).
bubble([H1,H2|T],[X1|T1],X):-
    hole(H1,H2,X1,X2),
    bubble([X2|T],T1,X).

bubblesort([],L,L).

bubblesort(L,L3,Sorted) :-
    bubble(L,L2,X),
    bubblesort(L2,[X|L3],Sorted).

forth_sort(L,L2) :- bubblesort(L,[],L2).
\end{minted}
\caption{Forth sorting sketch (\textbf{T6})}
\label{lst:forth-sort}
\end{lstfloat}

In Listing~\ref{lst:forth-sort}, there's a single neural predicate: \texttt{swap/3}. Its first two arguments are the numbers that are compared, the last argument is an indicator whether to swap or not. The \texttt{bubble/3} predicate performs a single step of bubble sort on its first argument using the \texttt{hole/4} predicate. The second argument is the resulting list after the bubble step, but without its last element, which is the third argument. The \texttt{bubblesort/3} predicate uses the \texttt{bubble/3} predicate, and recursively calls itself on the remaining list, adding the last element on each step to the front of the sorted list.

\begin{lstfloat}[ht]
\footnotesize
\begin{minted}[frame=lines]{Prolog}
permute(0,A,B,C,A,B,C).
permute(1,A,B,C,A,C,B).
permute(2,A,B,C,B,A,C).
permute(3,A,B,C,B,C,A).
permute(4,A,B,C,C,A,B).
permute(5,A,B,C,C,B,A).

swap(0,X,Y,X,Y).
swap(1,X,Y,Y,X).

operator(0,X,Y,Z) :- Z is X+Y.
operator(1,X,Y,Z) :- Z is X-Y.
operator(2,X,Y,Z) :- Z is X*Y.
operator(3,X,Y,Z) :- Y > 0, 0 =:= X mod Y,Z is X//Y.

nn(m_net1,[Repr],Y,[0,...,5])::net1(Repr,Y).
nn(m_net2,[Repr],Y,[0,...,3])::net2(Repr,Y).
nn(m_net3,[Repr],Y,[0,1])::net3(Repr,Y).
nn(m_net4,[Repr],Y,[0,...,3])::net4(Repr,Y).

wap(Text,X1,X2,X3,Out) :-
    net1(Text,Perm),
    permute(Perm,X1,X2,X3,N1,N2,N3),
    net2(Text,Op1),
    operator(Op1,N1,N2,Res1),
    net3(Text,Swap),
    swap(Swap,Res1,N3,X,Y),
    net4(Text,Op2),
    operator(Op2,X,Y,Out).
\end{minted}
\caption{Forth WAP sketch (\textbf{T7})}
\label{fig:forth}
\end{lstfloat}

In Listing~\ref{fig:forth}, there are four neural predicates: \texttt{net1/2} to \texttt{net4/2}. Their first argument is the input question, and the second argument are indicator variables for the choice of respectively: one of six permutations, one of 4 operations, swapping and one of 4 operations. These are implemented in the \texttt{permute/7}, \texttt{swap/5} and \texttt{operator/4} predicates. The \texttt{wap/5} predicate then sequences these steps to calculate the result.\\

\begin{lstfloat}[ht]
\footnotesize
\begin{minted}[frame=lines]{Prolog}
nn(net1, [X], Y, [heads, tails]) :: coin1(X,Y).
nn(net2, [X], Y, [heads, tails]) :: coin2(X,Y).

compare(X,X,same).
compare(X,Y,different) :- \+compare(X,Y,same).

coins(X,Comparison) :-
    coin1(X,C1),
    coin2(X,C2),
    compare(C1,C2,Comparison).
\end{minted}
\caption{The coins experiment (\textbf{T8})}
\label{lst:t8}
\end{lstfloat}

In Listing~\ref{lst:t8}, there are two neural predicates: \texttt{coin1/2} and \texttt{coin2/2}. Their input is the image of the two coins (e.g. Figure~\ref{fig:t8}). The output is heads or tails. The \texttt{coins/2} classifies both coins using these two predicates and then performs the comparison of the classes with the \texttt{compare/3} predicate.\\

\begin{lstfloat}[ht]
\footnotesize
\begin{minted}[frame=lines]{Prolog}
t(1/4)::house_rank(jack);t(1/4)::house_rank(queen);
    t(1/4)::house_rank(king);t(1/4)::house_rank(ace).
nn(net1,[X],Y,[jack,queen,king,ace]):: rank(X,Y).

hand(Cards,straight(low)) :-
    member(card(jack),Cards),
    member(card(queen),Cards),
    member(card(king),Cards).
hand(Cards,straight(high)) :-
    member(card(queen),Cards),
    member(card(king),Cards),
    member(card(ace),Cards).
hand([card(R), card(R), card(R)],threeofakind(R)).
hand(Cards,pair(R)) :-
    select(card(R),Cards,Cards2),
    member(card(R),Cards2).
hand(Cards,high(R)) :-
    member(card(R),Cards).

hand_rank(high(jack),0).
...
hand_rank(straight(high),13).

best_hand_rank(Cards,R) :-
    hand(Cards,H),
    hand_rank(H,R),
    \+(hand(Cards,H2),hand_rank(H2,R2),R2>R).

outcome(R1,R2,win) :- R1 > R2.
outcome(R1,R2,loss) :- R1 < R2.
outcome(R,R,draw).

cards(C1,C2,House,[card(R1), card(R2), House]) :-
    rank(C1,R1),
    rank(C2,R2).

game([C1,C2,C3,C4],House,Outcome) :-
    cards(C1,C2,House,Hand1),
    cards(C3,C4,House,Hand2),
    best_hand_rank(C1,R1),
    best_hand_rank(C2,R2),
    outcome(R1,R2,Outcome).

game(Cards,Outcome) :-
    house_rank(House),
    game(Cards,House,Outcome).
\end{minted}
\caption{The Poker experiment (\textbf{T9})}
\label{lst:t9}
\end{lstfloat}

\begin{figure}[t]
    \centering
    \includegraphics[width=0.2\linewidth]{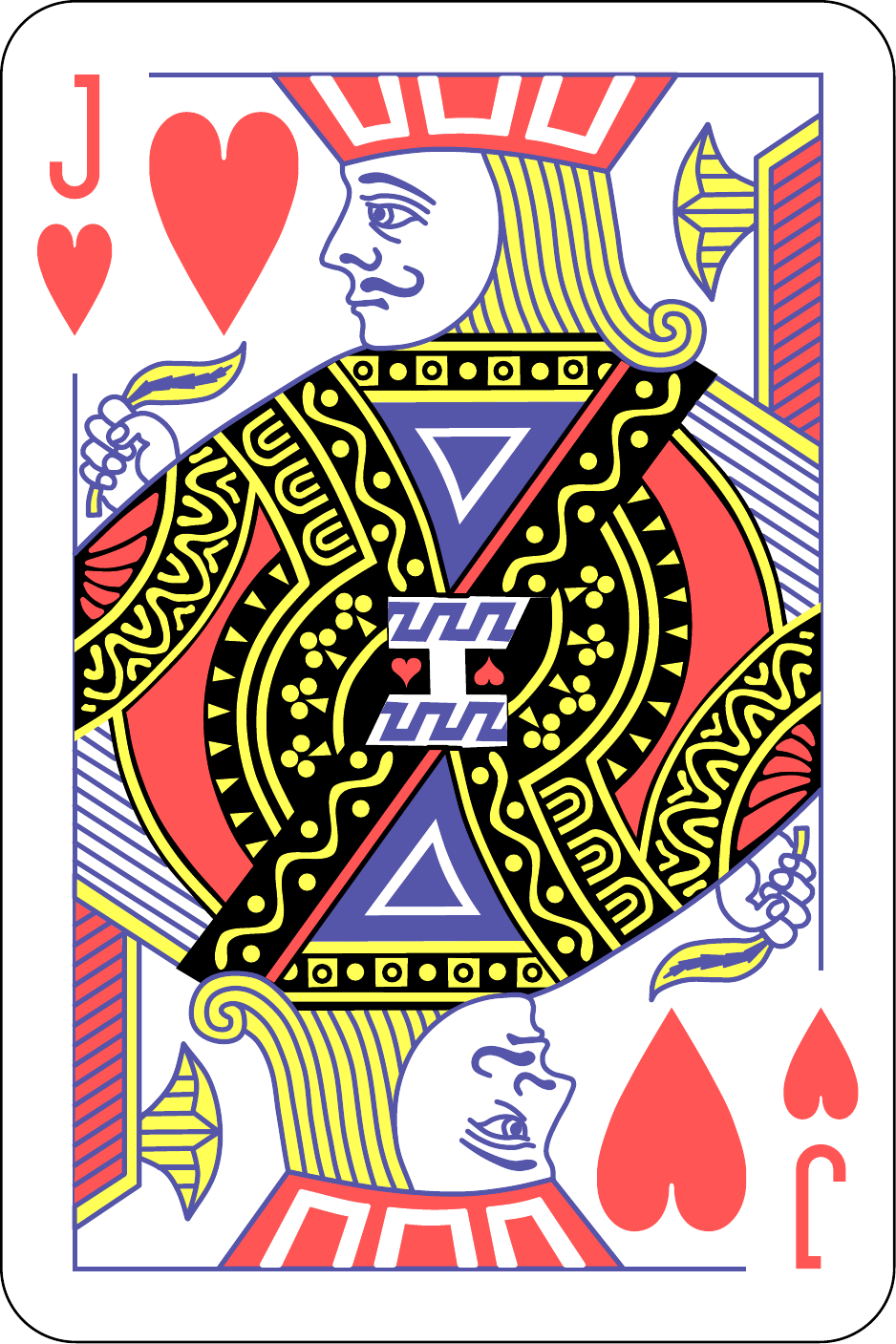}
    \includegraphics[width=0.2\linewidth]{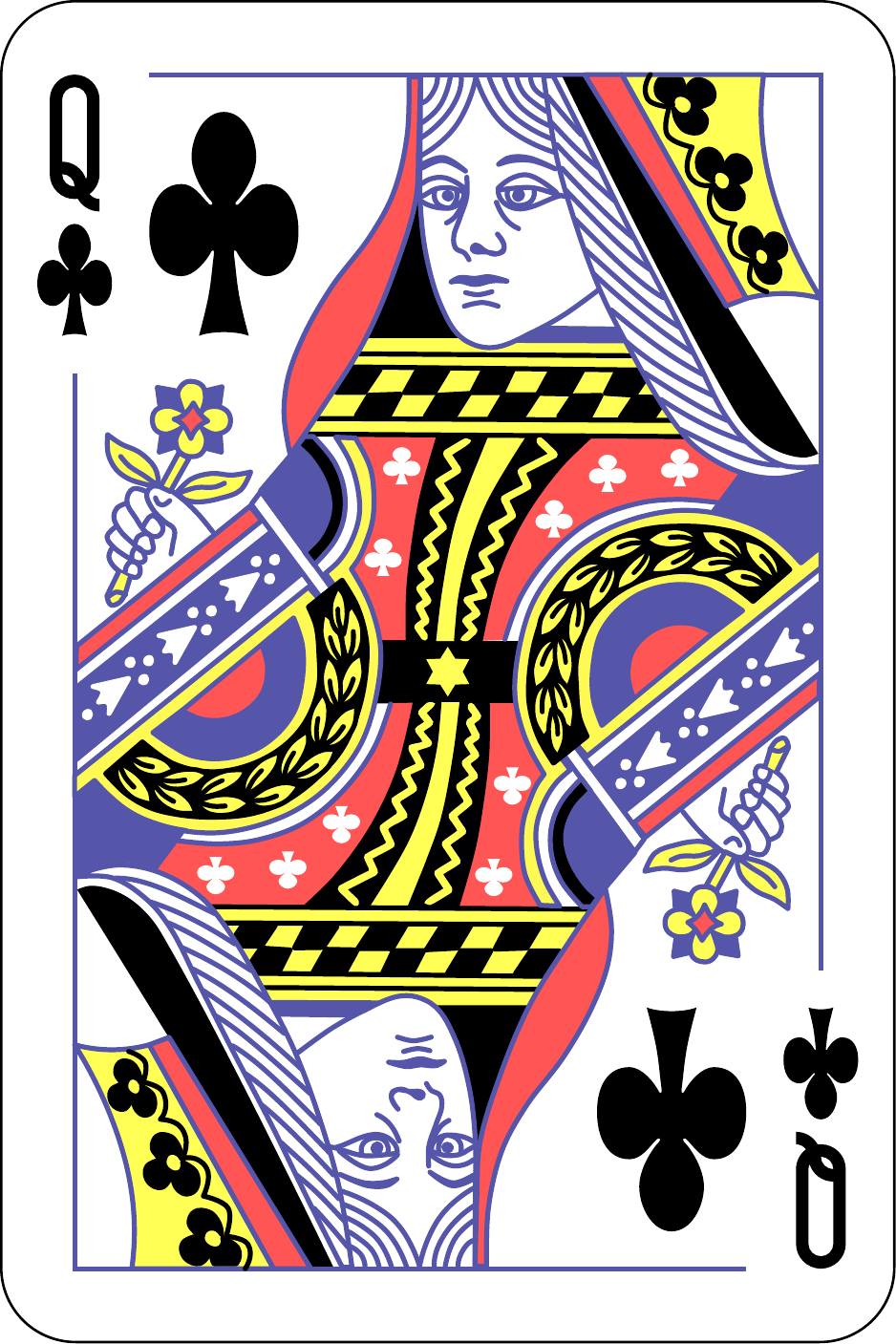}
    \includegraphics[width=0.2\linewidth]{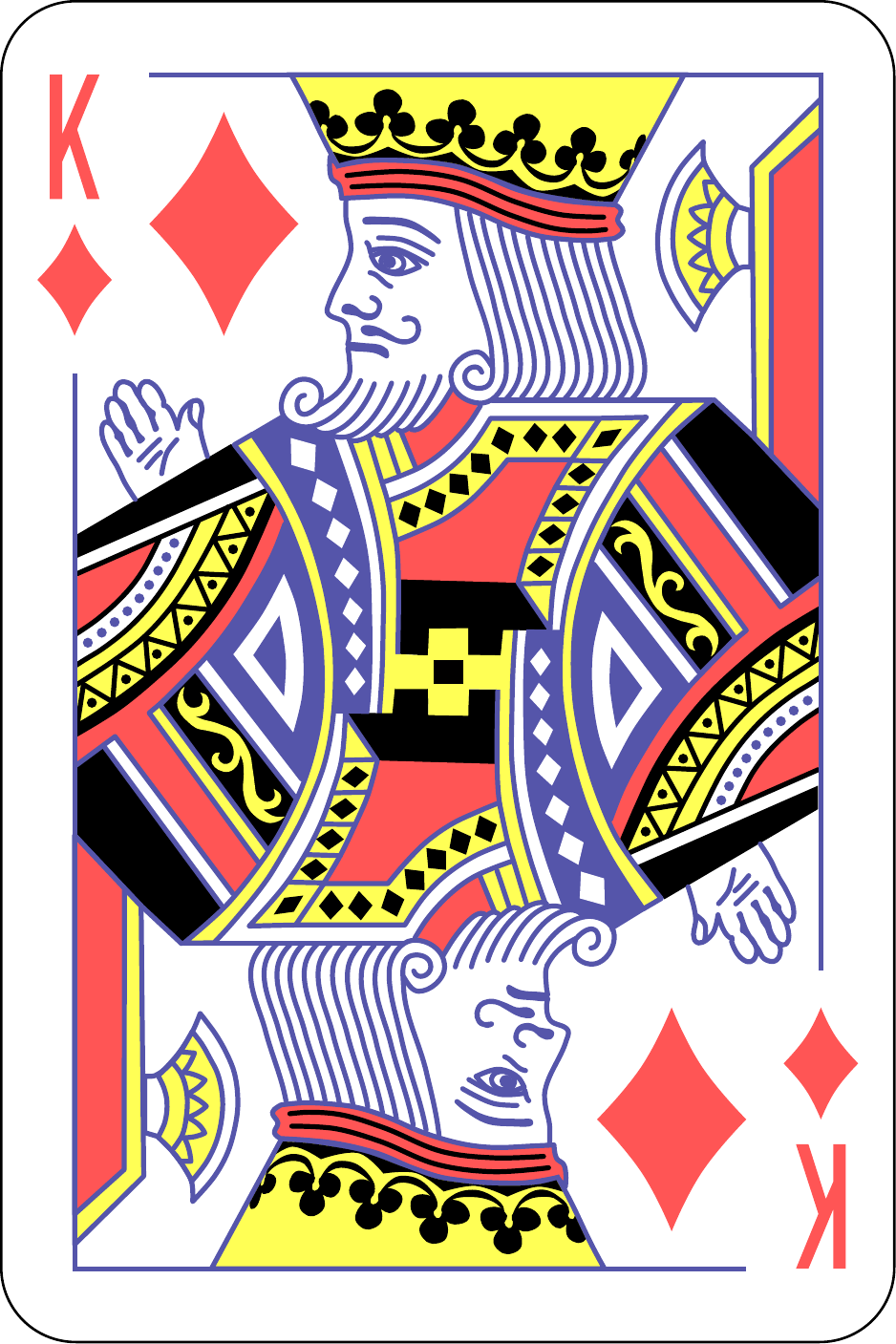}
    \includegraphics[width=0.2\linewidth]{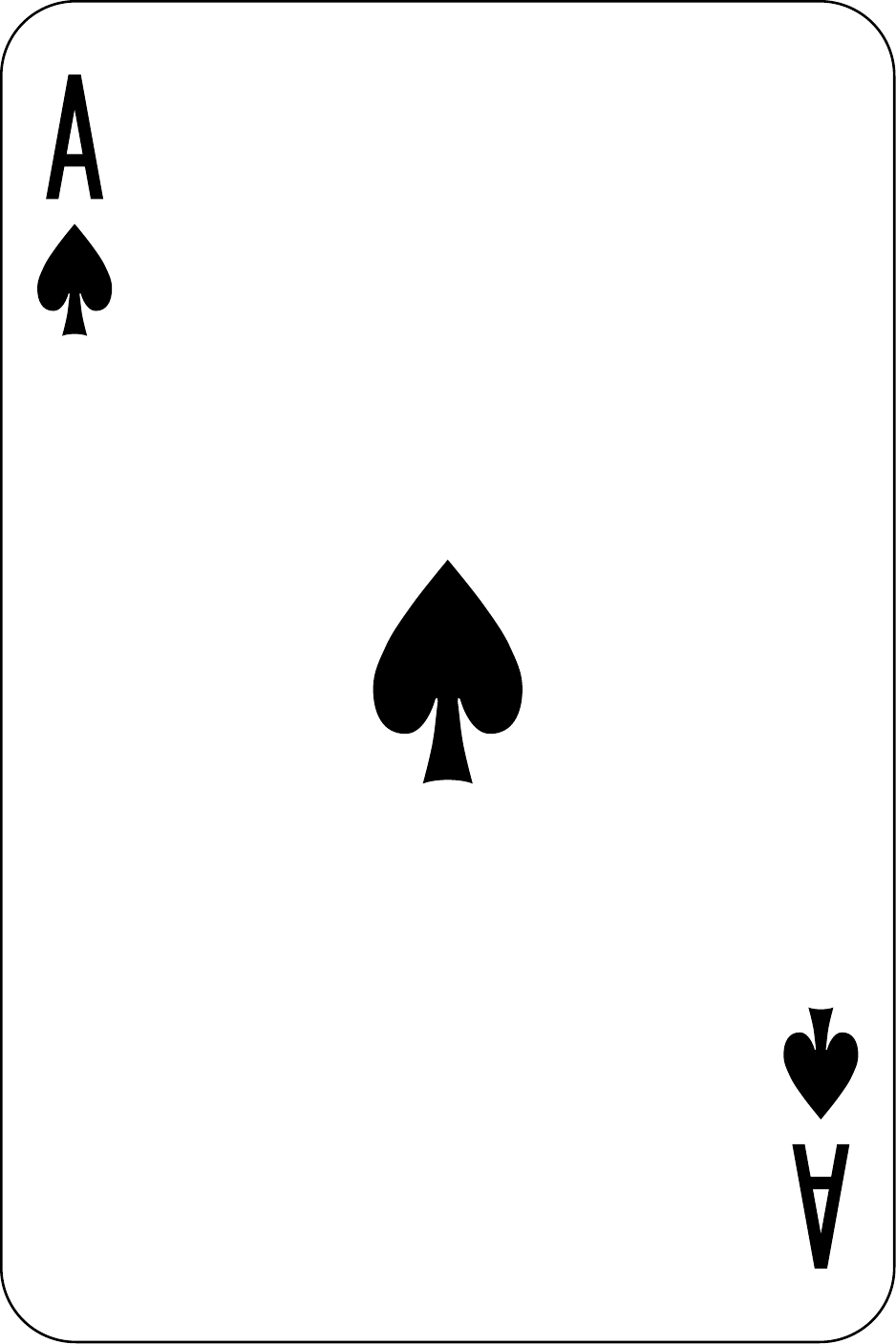}

    \caption{Examples of cards used as input for the Poker without perturbations(\textbf{T9}) experiment.}
    \label{fig:t9}
\end{figure}

In Listing~\ref{lst:t9}, there's a single neural predicate \texttt{rank/2} that takes as input the image of a card and classifies it as either a jack, queen, king or ace. There's also an AD with learnable parameters that represents the distribution of the unseen community card (\texttt{house\_rank/1}).  The \texttt{hand/2} predicate's first argument is a list of 3 cards. It unifies the output with any of the valid hands that these cards contain. The valid hands are: high card, pair (two cards have the same rank), three of a kind (three cards have the same rank), low straight (jack, queen king) and high straight(queen, king, ace).
Each hand is assigned a rank with the \texttt{hand\_rank/2} predicate. The \texttt{best\_hand\_rank/2} predicate takes as input a list of cards, and unifies the second argument with the highest hand rank that is possible with the three given cards. The \texttt{outcome/3} predicate determines the outcome by comparing the two ranks of the best hand.
The \texttt{game/3} predicate's first argument is a list of the 4 input images. It's second input is the labeled community card. It classifies the cards using the neural predicates, determines the best rank, and then unifies the last argument with the outcome.
The \texttt{game/2} determines the community card from the learned distribution \texttt{house\_rank/1}, and then determines the outcome using the \texttt{game/3} predicate.
The \texttt{member/2} and \texttt{select/3} predicates are predicates from the \textit{lists} library. \texttt{member/2} is true if it's second argument is a list and the first argument appears in that list. \texttt{select/3} is true if it's second argument is a list and the first argument appears in that list. It also unifies the last argument with the list that is the same as it's second argument, but with the first argument removed.
\end{document}